\documentclass[sn-mathphys,iicol]{sn-jnl}

\usepackage{microtype}
\usepackage{graphicx}
\usepackage{subfigure}
\usepackage{balance}
\usepackage{booktabs} 
\usepackage{multirow}
\usepackage{makecell}
\usepackage{adjustbox}
\usepackage{color}
\usepackage{graphicx}%
\usepackage{multirow}%
\usepackage{amsmath,amssymb,amsfonts}%
\usepackage{amsthm}%
\usepackage{mathrsfs}%
\usepackage[title]{appendix}%
\usepackage{xcolor}%
\usepackage{textcomp}%
\usepackage{manyfoot}%
\usepackage{booktabs}%
\usepackage{algorithm}%
\usepackage{algorithmicx}%
\usepackage{algpseudocode}%
\usepackage{listings}%
\usepackage{hyperref}
\usepackage{bbding}

\usepackage{url}
\usepackage{yfonts}
\usepackage{amssymb}
\usepackage{amsbsy}
\usepackage{amsmath}
\hypersetup{
    bookmarks=true,         
    unicode=false,          
    pdftoolbar=true,        
    pdfmenubar=true,        
    pdffitwindow=false,     
    pdftitle={Certificate},    
    pdfauthor={Dr. Harish Kumar},     
    pdfsubject={TEQIP certificates},   
    pdfcreator={Dr. Harish Kumar},   
    pdfproducer={},  
    pdfkeywords={Certificates,} {TEQIP} {Participation}, 
    pdfnewwindow=true,      
    colorlinks=false,       
    linkcolor=red,          
    citecolor=green,        
    filecolor=magenta,      
    urlcolor=cyan           
}

\theoremstyle{thmstyleone}%
\newtheorem{theorem}{Theorem}
\theoremstyle{thmstyletwo}%
\newtheorem{remark}{Remark}%

\theoremstyle{thmstylethree}%

\begin{document}
\title[Domain Generalization with Small Data]{Domain Generalization with Small Data}


\author[1]{\fnm{Kecheng} \sur{Chen}}\email{kechechen3-c@my.cityu.edu.hk}

\author[2]{\fnm{Elena} \sur{Gal}}\email{elena.gal@maths.ox.ac.uk}

\author[1]{\fnm{Hong} \sur{Yan}}\email{h.yan@cityu.edu.hk}
\author*[1]{\fnm{Haoliang} \sur{Li}}\email{haoliang.li@cityu.edu.hk}

\affil*[1]{\orgdiv{Department of Electrical Engineering and Center for Intelligent Multidimensional Data Analysis}, \orgname{City University of Hong Kong}, \orgaddress{\street{Tat Chee Avenue}, \city{Kowloon}, \state{Hong Kong SAR}}}
\affil[2]{\orgdiv{Department of Engineering}, \orgname{University of Oxford}, \orgaddress{\street{Old Road Campus Research Building, Roosevelt Drive}, \city{Oxford OX3}, \postcode{7DQ}, \country{UK}}}


\abstract{In this work, we propose to 
tackle the problem of domain generalization in the context of \textit{insufficient samples}. 
Instead of extracting latent feature embeddings based on deterministic
models, we propose to learn a domain-invariant representation based on the probabilistic framework by mapping each data point into probabilistic embeddings. Specifically, we first extend empirical maximum mean discrepancy (MMD) to a novel probabilistic MMD that can measure the discrepancy between mixture distributions (\textit{i.e.}, source domains) consisting of a series of latent distributions rather than latent points. Moreover, instead of imposing the contrastive semantic alignment (CSA) loss based on pairs of latent points, a novel probabilistic CSA loss 
encourages positive probabilistic embedding pairs to be closer while pulling other negative ones apart. Benefiting from the learned representation captured by probabilistic models, our proposed method can marriage the measurement on the \textit{distribution over distributions} (\textit{i.e.}, the global perspective alignment) and the distribution-based contrastive semantic alignment (\textit{i.e.}, the local perspective alignment). Extensive experimental results on three challenging medical datasets show the effectiveness of our proposed method in the context of insufficient data compared with state-of-the-art methods.}




\maketitle

\section{Introduction}
\begin{figure*}[!h]
    \centering
    \includegraphics[width=0.75\textwidth]{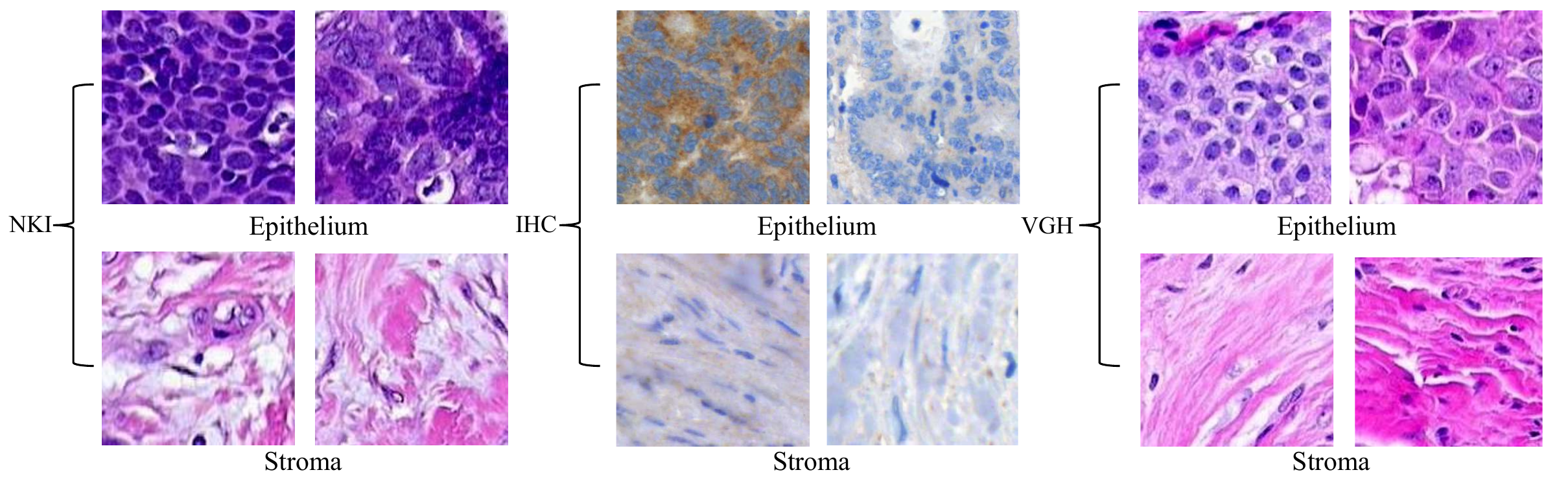}
    \caption{Histopathological image examples of breast cancer tissue from three different healthcare institutes, including NKI with 626 images, IHC with 645 images, and VGH with 1324 images. There are two different tissue types, including epithelium and stroma. Obvious domain gaps (\textit{e.g.}, the density of tissue and the staining color) can be observed. }
    \label{figure_es_in_intro}
\end{figure*}

Nowadays, we have witnessed a lot of successes with the application of machine learning techniques in a variety of tasks related to computer vision \citep{li2022comprehensive,zaidi2022survey} and natural language processing \citep{mridha2022study}.
Despite many achievements so far, the widely-adopted assumption for most existing methods, \textit{i.e.}, the data are identically and independently distributed in training and testing, may not always hold in actual applications \citep{zhou2022domain,liu2022deep}. In the real-world scenario, it is quite common that the distributions between training and testing data may be different, owing to changed environments. For example,  acquired histopathological images of breast cancer from different healthcare centers exhibit significant domain gaps  (\textit{a.k.a.}, domain shift, see Figure \ref{figure_es_in_intro} for more detail) caused by differences in device vendors and staining methods, which may lead to the catastrophic deterioration of the performance \citep{qi2020curriculum}. To address this issue, \textit{domain generalization} (DG) is developed to learn a model from multiple related yet different domains (\textit{a.k.a.}, source domains) that is able to generalize well on unseen testing domain (\textit{a.k.a.}, target domain).

Recently, researchers proposed several domain generalization approaches,
such as data augmentation with randomization \citep{yue2019domain}, data generalization with stylization \citep{pmlr-v97-verma19a,zhou2021domain}, meta learning-based training schemes \citep{li2018learning,kim2021self}, among which representation learning-based methods 
are one of the most popular ones. 
These representation learning-based methods \citep{balaji2019normalized} aim to learn domain-invariant feature representation. To be specific, 
if the discrepancy between source domains in feature space
can be minimized, the model is expected to generalize better on unseen target domain, due to learned domain-invariant and transferable feature
representation \citep{ben2006analysis}. For instance, 
an classical contrastive semantic alignment (CSA) loss proposed by \cite{motiian2017unified} was to encourage positive sample pairs (with same label) from different domains closer while pulling other
negative pairs (with different labels) apart. \cite{dou2019domain} introduced the CSA loss which jointly considers \textit{local class alignment loss} (for point-wise domain alignment) and \textit{global class alignment loss} (for distribution-wise alignment).

Despite the progress so far,  a reliable contrastive semantic loss with point-wise (or local) perspective usually requires sufficient
samples on source domains such that diverse sample-to-sample pairs can be constructed \citep{sohn2016improved,khosla2020supervised}. For example, \citet{khosla2020supervised} proposed a supervised contrastive semantic loss with a considerable volume of batch size on large-scale datasets such that decent performance can be guaranteed.
\citet{yao2022pcl} also emphasized the importance of  
the number of sample-to-sample pairs influenced by data sizes for contrastive-based loss on DG problem. On the other hand, in the eye of distribution-wise (\textit{a.k.a.}, global) alignment between domains \citep{dou2019domain}, a consistent distribution measurement (\textit{e.g.}, Kullback–Leibler (KL) divergence) theoretically relies on sufficient samples for the distribution estimation as discussed by \citep{bu2018estimation}. However, these sufficient samples from multiple source domains may not always be \textit{available} or \textit{accessible} in the real world. For example, for the medical imaging data, insufficient sample scenarios either exist in \textit{all source domains} (\textit{e.g.}, rare diseases inherently have a small volume of data from all healthcare centers \citep{lee2022deep}) or in \textit{some source domains} (\textit{e.g.}, some specific domains have significantly smaller sample sizes than others, resulting from the differences of the ethnicity \citep{johnson2022does}, the demography \citep{gurdasani2019genomics}, and the privacy-preserving regulation \citep{can2021privacy}). 
It is therefore necessary to develop reliable and effective semantic alignments from both local and global perspectives in the context of insufficient samples (\textit{a.k.a.}, small-data scenario) based on the source domains, in order to achieve better domain-invariant representations.

In this paper, we propose to learn domain-invariant representation from multiple source domains to tackle the domain generalization problem in the context of \textit{insufficient samples}. 
Instead of extracting latent embeddings (\textit{i.e.}, latent points) based on deterministic models (\textit{e.g.}, convolutional neural networks, CNNs), we propose to leverage a probabilistic framework endowed by variational Bayesian inference to map each data point into probabilistic embeddings (\textit{i.e.}, the latent distribution) for domain generalization. Specifically, by following the domain-invariant learning from global (distribution-wise) perspective, we propose to extend empirical maximum mean discrepancy (MMD) to a novel probabilistic MMD (P-MMD) that can empirically measure the discrepancy between mixture distributions (\textit{a.k.a.}, \textit{distributions over distributions}), consisted of a serial of latent distributions rather than latent points. From a local perspective, instead of imposing the CSA loss based on pairs of latent points, a novel probabilistic contrastive semantic alignment (P-CSA) loss
with kernel mean embedding is proposed to encourage positive probabilistic embedding pairs closer while pulling other negative ones apart.  Extensive experimental results on three challenging medical
imaging classification tasks, including 
epithelium stroma classification on insufficient histopathological images, skin lesion classification, and spinal cord gray matter segmentation, show that our proposed method can achieve better cross-domain performance in the context of insufficient data compared with state-of-the-art methods.
\section{Related Works}
\subsection{Domain Generalization with Medical Images}
Existing DG methods can be generally categorized into three different streams, namely data augmentation/generation \citep{yue2019domain,graves2011practical,zhou2021domain}, meta-learning \citep{li2018learning,kim2021self} and feature representation learning \citep{li2018domain,gong2019dlow,xiao2021bit}. Among these methods, feature representation learning, which aims to explore invariant feature information that can be shared across domains, demonstrates to be a widely adopted method for the problem of DG. 
For the feature representation learning-based DG method, \citep{li2018domain} proposed to conduct multi-domain alignment in latent space via a multi-domain MMD distance. \citet{gong2019dlow} leveraged adversarial training to eliminate the domain discrepancy such that domain-invariant representation can be learned in a manifold space. Due to the varieties of imaging protocol (\textit{e.g.}, the choice of image solution for MRI image), device vendors (\textit{e.g.}, Philips or Siemens CT scanners), and patient populations (the race and age group), the acquired imaging data from different medical sites may exist significant domain shift problem \citep{liu2021feddg}. \citet{dou2019domain} proposed a meta-learning framework to perform local and global semantic alignment for medical image classification. A similar design is also adopted by \citet{li2022domain} for tissue image classification. \citet{qi2020curriculum} utilized the curriculum learning scheme to transfer the knowledge for histopathological images classification.  \citet{li2020domain} combined the data augmentation and domain alignment to achieve decent performance on multiple medical data classification tasks. However, these methods may not focus on learning domain-invariant representation on \textit{insufficient samples} from source domains. 

\subsection{Probabilistic Neural Networks}
Compared with deterministic models, probabilistic neural networks turns to learn a distribution over model parameters, which can integrate the uncertainty in predictive modeling \citep{kingma2015variational,gal2016dropout}. When the data is insufficient, probabilistic models usually can achieve better generalized performance due to its probabilistic property (as an implicit regularization) \citep{blundell2015weight}. In the context of insufficient samples, Bayesian neural network \citep{neal2012bayesian} (BNN) with variational inference, a representative probabilistic model, not only can improve predictive accuracy as a classifier \citep{wilson2020bayesian}, but also can 
build up the quality of low-dimensional embeddings of insufficient data \citep{mallick2021deep}, which is a crucial motivation for this paper. Meanwhile, modern analytical approximation techniques (\textit{e.g.}, Variational inference \citep{blei2017variational}, empirical Bayes \citep{krishnan2020specifying}) can efficiently infer the posterior distribution of model parameters with stochastic
gradient descent method, which can integrate BNN with deterministic DNN conveniently.   

In \citet{xiao2021bit}, the authors proposed to consider the uncertainty of a generalizable model based on BNN, where the distances of positive probabilistic embedding pairs and class distribution are minimized via KL measure. Despite the effectiveness, the dissimilar pairs (\textit{i.e.}, negative pairs) are ignored, which may not benefit feature representation learning. Moreover, they only focused on sample similarity while the distribution information is ignored. Instead, our proposed method comprehensively considers both positive and negative probabilistic embedding pairs via a novel distribution-based contrastive semantic loss. Last but not the least, our proposed method highlights the benefit of the BNN for building up the quality of latent embeddings under insufficient sample scenarios.

\subsection{Probabilistic Embedding} Compared with deterministic point embeddings, probabilistic embeddings aim to characterize the data with a distribution. Due to its high robustness and effective representation \citep{nguyen2017mixture}, probabilistic representation has been applied to several fields, such as video representation learning \citep{park2022probabilistic}, image representation learning \citep{oh2018modeling}, face recognition \citep{shi2019probabilistic,chang2020data}, speaker diarization \citep{silnova2020probabilistic}, and human pose estimation \citep{sun2020view}. Recently, some researchers further leveraged the probabilistic embeddings to bridge the gap between data modalities \citep{chun2021probabilistic,neculai2022probabilistic,chun2023improved}. For example, \citet{chun2021probabilistic} found that the probabilistic representation can lead to a richer embedding space for the challengeable relation reasoning between the images and their captions. These probabilistic embedding-based approaches either inherit the inherent distribution property of data (\textit{e.g.}, the multiple frames of a video) or tackle the one-to-many correspondences through distributional representation. Instead, our proposed method imposes the Bayesian neural network to generate probabilistic embedding. As such, the representative capacity of the data in the small-data regime can be enhanced. Moreover, we devise a novel probabilistic MMD to measure the discrepancy between mixture probabilistic embeddings for domain-invariant learning.
\begin{figure*}[!t]
    \centering
\includegraphics[width=0.85\textwidth]{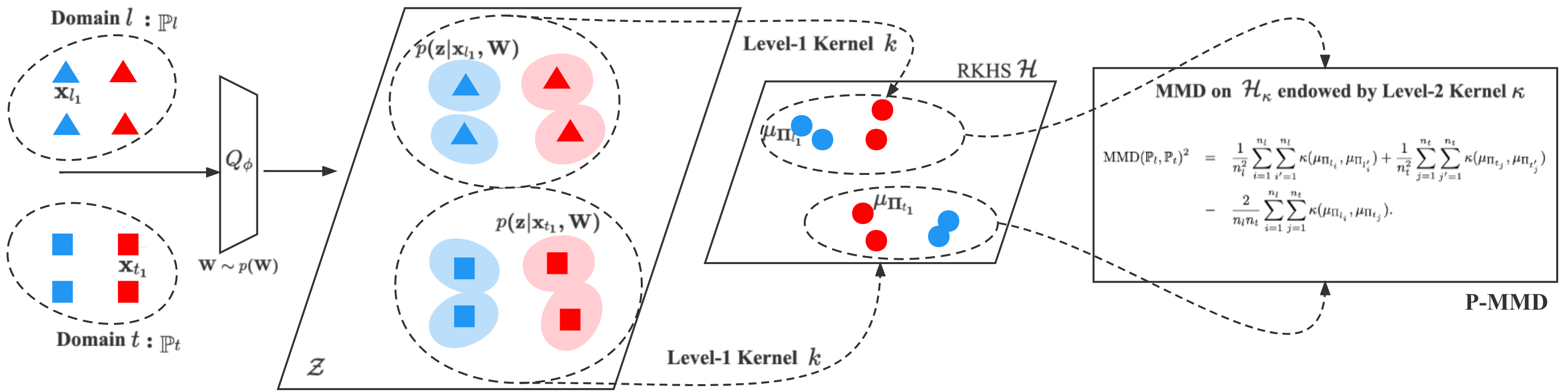}
    \caption{A visualized computational process for probabilistic MMD (P-MMD) on two source domains. The same color for samples in different domains denotes the same label.}
    \label{p-mmd_visualized}
    
\end{figure*}
\section{Methodology}
\textbf{Preliminary. }
Assume that there are $K$ domains from different collected environments.
The samples in each domain can be represented as $\mathbf{X}_{l} = \{\mathbf{x}_{l_{1}},\cdots,\mathbf{x}_{l_{n_{l}}}\}$, where $l \in \mathbb{N}^{+}: \{1,\cdots,K\}$, $\mathbf{x}_{l_{i}} \in \mathbb{R}^{d \times 1}$ denotes a sample with the $d$ dimension vector in the $l$-th domain. $n_{l}$ is the total number of samples in the $l$-th domain. The corresponding labels of samples $\mathbf{X}_{l}$ in each domain can be denoted as $\mathbf{Y}_{l} = \{\mathbf{y}_{l_{1}},\cdots,\mathbf{y}_{l_{n_{l}}}\}$, where $\mathbf{y}_{l_{i}} \in \mathbb{R}^{m \times 1}$ is the form of one-hot encoding with $m$ classes in total. For the setting of domain generalization, the source domain data represented as $\{\mathbf{X}_{l}^{S},\mathbf{Y}_{l}^{S}\}_{l=1}^{K}$, can be available in the training phase only, whereas the target domain data, denoted by $\mathbf{X}^{T}$, are only seen in test phase. 

\textbf{Overall.} We provide a framework that can learn better domain-invariant representation when there is insufficient source domain data. The probabilistic neural network is imposed to enable high-quality and powerful feature representation in the context of insufficient samples. To effectively perform global perspective alignment, a novel probabilistic MMD is proposed to empirically measure the discrepancy between distributions over distributions based on 
reproducing kernel Hilbert space. We also propose a probabilistic contrastive semantic alignment to adapt probabilistic embeddings with local perspective. The details of our proposed method are discussed as below.

\textbf{Probabilistic Embedding of Insufficient Data. }
Compared with deterministic models, the probabilistic models can learn a distribution over model weights, which has shown a better capacity to represent latent embeddings under insufficient sample scenario \citep{mallick2021deep}. In this work, Bayesian neural network (BNN) \citep{blei2017variational} is utilized to extract the low-dimensional embeddings from high-dimensional inputs. By feeding the inputs into BNN with a parameter $\mathbf{W} \sim p(\mathbf{W)}$ , the samples $\mathbf{X}_{l} = \{\mathbf{x}_{l_{1}},\cdots,\mathbf{x}_{l_{n_{l}}}\}$ of each domain can be represented by a set of probabilistic embeddings (\textit{i.e.}, latent distributions), 
\textit{i.e.}, $p(\mathbf{Z}|\mathbf{X}_{l}) = \{p(\mathbf{z}|\mathbf{x}_{l_{1}},\mathbf{W}),\cdots,p(\mathbf{z}|\mathbf{x}_{l_{n_{l}}},\mathbf{W})\}$ where $\mathbf{W} \sim p(\mathbf{W)}$ is sampled stochastically.  The variational inference is used to approximate the posterior distribution of $\mathbf{W}$ with the evidence lower bound (ELBO) (more details can be found in \hyperref[bnn]{Appendix A.1}). By using Monte Carlo (MC) estimators with $T$
stochastic sampling operations from $\mathbf{W}$, the predictive distribution of each $p(\mathbf{z}|\mathbf{x})$ can be an unbiased approximation. 
\subsection{Distribution Alignment via Probabilistic Maximum Mean Discrepancy}
In this section,  we introduce an approach to learning domain-invariant representation from a global perspective by minimizing the discrepancy among domains. Among various distribution distance metrics, Maximum Mean Discrepancy (MMD) is widely adopted \citep{long2017deep,li2018domain} which aims to measure the distance between two probability distributions in a non-parametric manner.
Specifically, assume that latent embeddings $\mathbf{Z}_{l}$=$\{\mathbf{z}_{l_{1}},\cdots,\mathbf{z}_{l_{n_{l}}}\}$ and $\mathbf{Z}_{t}$=$\{\mathbf{z}_{t_{1}},\cdots,\mathbf{z}_{t_{n_{t}}}\}$ are drawn from two unknown distributions $\mathbb{P}_{l}$ and $\mathbb{P}_{t}$. The probability measure $\mathbb{P}$ can be mapped into a reproducing kernel Hilbert space (RKHS) $\mathcal{H}$ as a element by setting,
\begin{equation}\label{kernel_embedding}
    \mu_{\mathbb{P}}:= \mathbb{E}_{\mathbf{z}\sim \mathbb{P}}[\phi(\mathbf{z})]= \int_{\mathcal{Z}}k(\mathbf{z},\cdot)d\mathbb{P}= \mathbb{E}_{\mathbf{z}\sim \mathbb{P}}[k(\mathbf{z},\cdot)],
\end{equation}
where a reproducing kernel $k: \mathcal{X} \times \mathcal{X} \rightarrow \mathbb{R}$ and corresponding feature map $\phi: \mathcal{X} \rightarrow \mathcal{H}$ are defined. Let the kernel $k$ is characteristic such that the map $\mu:\mathbb{P}\rightarrow \mu_{\mathbb{P}}$ is injective. In this case the MMD can be defined as the distance $\lVert\mu_{\mathbb{P}_l}-\mu_{\mathbb{P}_k}\rVert_{\mathcal{H}}$ in $\mathcal{H}$ between mean embeddings and it can be used as a measure of distance between the distributions $\mathbb{P}_{l}$ and $\mathbb{P}_{t}$ \citep{borgwardt2006integrating, gretton2012kernel}. The explicit computation of MMD can be derived by unbiased empirical estimation of mean map \citep{gretton2012kernel}, \textit{i.e.},
\begin{equation}\label{mmd}
\begin{aligned}
&\operatorname{MMD}\left(\mathbb{P}_{l}, \mathbb{P}_{t}\right)^{2}= \|\frac{1}{n_{l}} \sum_{i=1}^{n_{l}} \phi\left(\mathbf{z}_{l_{i}}\right)-\frac{1}{n_{t}} \sum_{j=1}^{n_{t}} \phi\left(\mathbf{z}_{t_{j}}\right)\|_{\mathcal{H}}^{2}
\end{aligned}
\end{equation}
The idea of using MMD for domain generalization has been explored in several works (\textit{e.g.}, \citep{li2018domain,hu2020domain}). 

In the probabilistic framework, instead of the individual latent embeddings $\mathbf{z}_{l_{1},\ldots}$, we have latent probabilistic embeddings $\Pi_{l_{1}}:= p(\mathbf{z}|\mathbf{x}_{l_{1}},\mathbf{W}),\ldots$. For a source domain $D_l$, we have the associated \textit{distribution over distributions} $\mathbb{P}_{l} = \{\Pi_{l_{1}},\cdots,\Pi_{l_{n_{l}}}\}$. For this scenario, we propose to extend the existing \textit{point-based} empirical MMD estimate to a \textit{distribution-based} empirical probability MMD (P-MMD) estimate. P-MMD utilizes empirical estimation by kernels on distributions to measure the discrepancy between mixture distributions $\mathbb{P}_{l}$ and $\mathbb{P}_{t}$ under the probabilistic framework. 

Specifically, we first represent latent probabilistic embeddings as elements in RKHS $\mathcal{H}_k$ using the kernel $k$, that we coin a \emph{level-1} kernel in the sequel, \textit{e.g.}, $\mu_{\Pi_{l_{1}}} := \mathbb{E}_{\mathbf{z}\sim \Pi_{l_{1}}}[\phi(\mathbf{z})]=\mathbb{E}_{\mathbf{z}\sim \Pi_{l_{1}}}[k(\mathbf{z},\cdot)]$, which is an analog to the Eq. (\ref{kernel_embedding}). The kernel mean embedding $\mu_{\Pi_{l_{1}}}$ can be regarded as a new feature map for a variety of tasks \citep{yoshikawa2014latent}. 
Here, to enable \emph{non-linear} learning on distributions, we introduce a \emph{level-2} kernel $K$ \citep{muandet2012learning}. Consider a level-1 kernel $\kappa$ on  $\mathcal{H}$ and its reproducing kernel Hilbert space (RKHS) $\mathcal{H}_{\kappa}$. Define $K$ as
\begin{equation}
     K(\Pi_{l_{i}}, \Pi_{t_{j}}) = 
     \kappa(\mu_{\Pi_{l_{i}}},\mu_{\Pi_{t_{j}}}) = \langle \psi(\mu_{\Pi_{l_{i}}}),\psi(\mu_{\Pi_{t_{j}}})\rangle_{\mathcal{H}_{\kappa}},
\end{equation}
where $K$ and its explicit form on kernel mean embeddings $\kappa$ are p.d. kernels 
\citep{berlinet2011reproducing}. We define a novel \textit{probabilistic MMD} (P-MMD) empirical estimation method using the \emph{level-2} kernel $K$:
\begin{small}
    \begin{align}
\label{p-mmd-full}
    &\operatorname{P-MMD}(\mathbb{P}_{l}, \mathbb{P}_{t})^{2}  =
    \|\frac{1}{n_{l}} \sum_{i=1}^{n_{l}} \psi(\mu_{\Pi_{l_{i}}})-\frac{1}{n_{t}} \sum_{j=1}^{n_{t}} \psi(\mu_{\Pi_{t_{j}}})\|_{\mathcal{H_{\kappa}}}^{2}\nonumber\\&=
    \frac{1}{n_{l}^{2}} \sum_{i=1}^{n_{l}} \sum_{i^{\prime}=1}^{n_{l}} K(\Pi_{l_{i}}, \Pi_{l_{i}^{\prime}}) + \frac{1}{n_{t}^{2}} \sum_{j=1}^{n_{t}} \sum_{j^{\prime}=1}^{n_{t}} K(\Pi_{t_{j}}, \Pi_{t_{j}^{\prime}})\nonumber 
\\&-\frac{2}{n_{l} n_{t}} \sum_{i=1}^{n_{l}} \sum_{j=1}^{n_{t}} K(\Pi_{l_{i}}, \Pi_{t_{j}}).
\end{align}
\end{small}

In this work, the level-1 and level-2 kernels, $k$ and $K$, are both Gaussian RBF kernel due to its impressive performance on a limited amount of distribution data \citep{muandet2012learning}. Namely, $K = K_{Gau}(\Pi_{l_{i}}, \Pi_{t_{j}}) = \kappa(\mu_{\Pi_{l_{i}}}, \mu_{\Pi_{t_{j}}})$ can be represented as 
\begin{small}
\begin{align}
\label{level2}
  & \kappa(\mu_{\Pi_{l_{i}}}, \mu_{\Pi_{t_{j}}}) =  \exp (-\frac{\lambda}{2}\|\mu_{\Pi_{l_{i}}} - \mu_{\Pi_{t_{j}}}\|_{\mathcal{H}_{\kappa}}^{2}) \nonumber \\ &= \exp(-\frac{\lambda}{2}(\langle \mu_{\Pi_{l_{i}}},\mu_{\Pi_{l_{i}}}\rangle_{\mathcal{H}_{\kappa}})- 2\langle \mu_{\Pi_{l_{i}}},\mu_{\Pi_{t_{j}}}\rangle_{\mathcal{H}_{\kappa}} + \nonumber\\ & \langle \mu_{\Pi_{t_{j}}},\mu_{\Pi_{t_{j}}}\rangle_{\mathcal{H}_{\kappa}})) \nonumber \\ &= \exp (-\frac{\lambda}{2}(\frac{1}{m_{l}^{2}} \sum_{i=1}^{m_{l}} \sum_{i^{\prime}=1}^{m_{l}}  k(\mathbf{z}_{l_{i}},\mathbf{z}_{l_{i}^{\prime}}) \nonumber  \\
    & -   \frac{2}{m_{l} m_{t}} \sum_{i=1}^{m_{l}} \sum_{j=1}^{m_{t}} k(\mathbf{z}_{l_{i}},\mathbf{z}_{t_{j}}))
     + \frac{1}{m_{t}^{2}} \sum_{j=1}^{m_{t}} \sum_{j^{\prime}=1}^{m_{t}}  k(\mathbf{z}_{t_{j}},\mathbf{z}_{t_{j}^{\prime}}),
\end{align}
\end{small}where $m_{l}$ and $m_{t}$ are determined by sampling times $T$. 
The kernel mean embedding using the level-1 kernel $k$ creates \emph{distributions} $\mu(\mathbb{P}_1),\dots,\mu(\mathbb{P}_N)$ represented by the samples $\{\mu_{\Pi_{l_1}},\ldots,\mu_{\Pi_{l_n}}\}$ for $l=1,\ldots,N$ respectively in the RKHS $\mathcal{H}_k$. The underlying strategy of P-MMD is to apply the classic MMD to these distributions (with respect to the kernel $\kappa$) . To access the effect that the minimization of P-MMD has on the original latent probability distributions across different domains, we recall the following: 
\begin{theorem}[\cite{muandet2012learning}]
\label{thm1}
Let $\mathbb{P}_1,\ldots,\mathbb{P}_N$  be probability distributions and $\hat{\mathbb{P}}:=\frac{1}{N}\sum_{i=1}^{N}\mathbb{P}_i$. Then the distributional variance given by $\frac{1}{N}\sum\lVert\mu_{\mathbb{P}_i}-\mu_{\hat{\mathbb{P}}}\rVert$ is 0 iff $\mathbb{P}_1=\mathbb{P}_2=\ldots=\mathbb{P}_N$.
\end{theorem}
\newtheorem{corollary}{\bf Corollary }[section]
\begin{corollary}[\cite{li2018domain}]
\label{cor1}
The upper bound of the distributional variance can be written as 
\begin{equation*}
\frac{1}{K^{2}} \sum_{1 \leq i,j \leq K}  \operatorname{MMD}(\mathbb{P}_{i}, \mathbb{P}_{j})^{2}.
\end{equation*}
\end{corollary}
In our setting Theorem \ref{thm1} and Corollary \ref{cor1} along with the fact that $k$ is a characteristic kernel imply the following:
\begin{corollary}
\label{cor2}
iff all moments of latent distributions $\Pi_{l}$ associated to points of domain $D_l$ for $l=1,\ldots,N$ are distributed identically across domains, $
  \frac{1}{K^{2}} \sum_{1 \leq i,j \leq K}  \operatorname{P-MMD}(\mathbb{P}_{i}, \mathbb{P}_{j})^{2}=0
$ holds.
\end{corollary}
Following Corollary \ref{cor2} we define the following loss function:
\begin{equation}
\label{loss_global}
    \mathcal{L}_{global} = \frac{1}{K^{2}} \sum_{1 \leq i,j \leq K}  \operatorname{P-MMD}(\mathbb{P}_{i}, \mathbb{P}_{j})^{2}.
\end{equation}

Corollary \ref{cor2} implies that as Eq. \ref{loss_global} tends to 0 so does the distance between the distributions of  means, variances and higher moments of the distributions $\Pi_l$
associated to points of different domains. 

\begin{remark}In section \ref{kme}, we compare the P-MMD approach to simply taking the mean (\textit{i.e.}, first moment) of latent probabilistic embeddings $\Pi_l$ \textit{i.e.} taking ${\Pi}_{l}\rightarrow \mathbf{m}_{\Pi_{l}}=\mathbb{E}_{\mathbf{x}\sim \Pi_{l}[\mathbf{x}]}$, and then minimizing the associated ``vanilla" MMD. Although this scheme is more efficient computationally over our proposed method, it discards most information about high-level statistics as discussed by \cite{muandet2017kernel}. We empirically verify that our approach has better performance across domains. The visualized computation of P-MMD is shown in Figure \ref{p-mmd_visualized}.
\end{remark}

Although we focus on the scenario of insufficient samples, the computational consumption from  Eqs. (\ref{p-mmd-full}) and (\ref{level2}) may be still prohibitive as the calculation of MMD distance between distributions can scale at least quadratically with the increasing of sample size (especially for image segmentation task), \textit{i.e.}, $O(n^{2})$ in a domain. 
Here, by following the \textit{linear statistic theory} of MMD, the unbiased estimate can be derived by drawing pairs from two domains with replacement, \textit{i.e.}, $\operatorname{P-MMD}(\mathbb{P}_{l}, \mathbb{P}_{t})^{2} \approx \frac{2}{n_{l}}\sum_{i=1}^{\frac{2}{n_{l}}}[K(\Pi_{l_{2i}}, \Pi_{l_{2i+1}^{\prime}})+K(\Pi_{t_{2i}}, \Pi_{t_{2i+1}^{\prime}})-K(\Pi_{l_{2i}}, \Pi_{t_{2i+1}})-K(\Pi_{l_{2i+1}}, \Pi_{t_{2i}})$, where assuming $n_{l}=n_{t}$ for simplicity. \cite{borgwardt2006integrating} gives proof about the unbiased property of the \textit{linear statistic} of MMD and shows that statistic power does not be sacrificed too much.
\begin{table*}[!h] 
\small
\centering
\renewcommand{\arraystretch}{0.8}
\caption{Experiment results of Epithelium Stroma Classification of
Histopathological Images.  Each column denotes a cross-domain task. For example, in the second column, we use IHC dataset as the target domain and the remaining datasets as the source domains. Note that all baseline methods adopt the SWAD method \citep{cha2021swad} for weight averaging. The baseline in the sixth row, namely SWAD, denotes the ERM training strategy with the SWAD method.} 
\label{table:pacs}
\begin{tabular}{lccccc}
\toprule[1pt]
Method &  & IHC & NKI & \multicolumn{1}{c}{VGH}  & Average (\%)  \\  
\hline
\multicolumn{2}{l}{DeepAll }& 73.29 $\pm$ 0.13 &  70.60 $\pm$ 0.15 & 79.56 $\pm$ 0.11 & 74.48\\
 \multicolumn{2}{l}{MASF \citep{dou2019domain} }& 80.45$\pm$ 0.10 &  \underline{76.10}$\pm$ 0.11 & 84.44$\pm$ 0.12 & 80.33\\ 
  \multicolumn{2}{l}{LDDG \citep{li2020domain} }& 81.19$\pm$ 0.23 &  73.27$\pm$ 0.25 & 82.58$\pm$ 0.23 & 79.01\\ 
 \multicolumn{2}{l}{KDDG \citep{wang2021embracing} }& 83.65$\pm$ 0.19 &  74.04$\pm$ 0.15 & 83.13$\pm$ 0.20 & 80.27\\ 
\multicolumn{2}{l}{SWAD \citep{cha2021swad}}  & 79.74$\pm$ 0.15 & 74.84$\pm$ 0.13 & 84.29$\pm$ 0.12 & 79.62\\
\multicolumn{2}{l}{BDIL \citep{xiao2021bit}}  & \underline{85.56}$\pm$ 0.12 & 71.89$\pm$ 0.14 & \underline{85.90}$\pm$ 0.18 & 81.05\\
\multicolumn{2}{l}{DNA \citep{chu2022dna}}& 83.93$\pm$ 0.18 & 73.94$\pm$ 0.15  & 85.57$\pm$ 0.17 & 81.14 \\
\multicolumn{2}{l}{DSU \citep{li2022uncertainty}} & 81.56$\pm$ 0.14 & 72.47$\pm$ 0.12  & 83.94$\pm$ 0.16 & 79.32 \\
\multicolumn{2}{l}{MIRO \citep{cha2022domain}} & 82.69$\pm$ 0.11 & 74.93 $\pm$ 0.13 & 84.63$\pm$ 0.11 & 80.80\\
 \multicolumn{2}{l}{Ours (in this paper)} & \textbf{88.82}$\pm$ 0.09 & \textbf{76.71}$\pm$ 0.10 & \textbf{86.92}$\pm$ 0.14 & \textbf{84.06} \\

\bottomrule[1pt]
\end{tabular}
\label{breast_exp}
\end{table*}

\subsection{Probabilistic Contrastive Semantic Alignment. }
To learn domain-invariant representation from a local perspective, a popular idea is to encourage positive pairs with same label closer together, while pulling other negative ones with different labels further apart \citep{motiian2017unified,dou2019domain}. These methods usually measure the Euclidean distance between samples in the embedding space. However, this scheme may not satisfy our probabilistic framework due to its probabilistic embeddings. 

To this end, we propose a probabilistic contrastive semantic alignment (P-CSA) loss that can utilize the empirical MMD to measure the discrepancy between  probabilistic embeddings. The proposed P-CSA loss $\mathcal{L}_{local}$ consists of two components, including the positive probabilistic contrastive loss and negative probabilistic contrastive loss. The former aims to minimize the distance between the intra-class distributions
from different domains, \textit{i.e.}, 
\begin{small}
    \begin{equation}
    \begin{aligned}
    &\mathcal{L}_{local}^{pos} =\frac{1}{2}\|\frac{1}{T} \sum_{i=1}^{T} \phi\left(M_{\Theta}(\mathbf{z}_{n_{i}})\right)-\frac{1}{T} \sum_{j=1}^{T} \phi\left(M_{\Theta}(\mathbf{z}_{q_{j}})\right)\|_{\mathcal{H}}^{2} ,
\end{aligned}
\end{equation}
\end{small}where $M_{\Theta}(\cdot)$ denotes the embedding network of metric learning, which will contribute to learn the distance between features better \citep{dou2019domain}. Note that $\mathbf{y}_{n}	= \mathbf{y}_{q}$ needs to be satisfied. Then, the negative probabilistic contrastive loss is denoted by 
\begin{small}
    \begin{align}
    &\mathcal{L}_{local}^{neg} =  \frac{1}{2} \max[0, \xi - \operatorname{MMD}(\Pi_{n},\Pi_{q})^{2}] = \frac{1}{2} \max [0, \xi \nonumber \\ & - \|\frac{1}{T} \sum_{i=1}^{T} \phi\left(M_{\Theta}(\mathbf{z}_{n_{i}})\right)-\frac{1}{T} \sum_{j=1}^{T} \phi\left(M_{\Theta}(\mathbf{z}_{q_{j}})\right)\|_{\mathcal{H}}^{2}],
\end{align}
\end{small}where $\xi$ is a distance margin that can guarantee an appropriate repulsion range. Note that  $\mathbf{y}_{n}	\neq \mathbf{y}_{q}$ needs to be satisfied.

\textbf{Model Training.} Our proposed framework consists of three modules, a BNN-based probabilistic extractor $Q_{\phi}$, a BNN-based classifier $C_{\omega}$, and a metric network $M_{\Theta}(\cdot)$. For the $Q_{\phi}$, we only add a Bayesian layer with ReLU layer on the bottom of a pretrained deterministic model (\textit{e.g.}, ResNet18 by removing fully-connected layers) by following \citet{xiao2021bit}. For the $C_{\omega}$, a Bayesian layer is also introduced to adapt the classification on insufficient sample better. More implement details of BNN can be found in \hyperref[bnn]{Appendix A.1}. The structure of $M_{\Theta}$ is the same as \citet{dou2019domain}. The  images $\mathcal{X}=\{\mathbf{x}_{l_{i}}\}$ conduct $T$ 
stochastic forward passes on the $Q_{\phi}$ and $C_{\omega}$ by MC sampling to obtain probabilistic predicts $\{\hat{y}_{l_{i}}^{j}\}_{j=1}^{T}$, where the outputs (\textit{i.e.}, probabilistic embeddings) of $Q_{\phi}$ serve as the inputs for the calculations of $\mathcal{L}_{global}$  and $\mathcal{L}_{local}$. The final predicts $\{\hat{y}_{l_{i}}^{j}\}$ are the expectation of  $\{\hat{y}_{l_{i}}^{j}\}_{j=1}^{T}$. The total objectives can be summarized as follows,
\begin{align}
    &\mathcal{L}_{total} = \sum_{l,i}\mathcal{L}_{c}(\hat{y}_{l_{i}},y_{l_{i}}) + \operatorname{KL}[q_{\theta}(Q_{\phi})\|p(Q_{\phi})] \nonumber \\ & + \operatorname{KL}[q_{\theta}(C_{\omega})\|p(C_{\omega})]+ \beta_{1}\mathcal{L}_{local} + \beta_{2}\mathcal{L}
_{global}.
\end{align}
\begin{table*}[!t]
\renewcommand{\arraystretch}{1.2}
\centering
\caption{Domain generalization results on skin lesion classification. Each column denotes a cross-domain task. For example, in the second column, we use DMF dataset as the target domain and the remaining datasets as the source domains. The best and second-best performance on each target domain are bolded and underlined, respectively.  Note that all baseline methods adopt the SWAD method \citep{cha2021swad} for weight averaging. The baseline in the sixth row, namely SWAD, denotes the ERM training strategy with the SWAD method.}
\begin{adjustbox}{width=\textwidth}
\begin{tabular}{llllllll}
\toprule[1.2pt]
Method  & \multicolumn{1}{c}{DMF}  & \multicolumn{1}{c}{D7P} & 
\multicolumn{1}{c}{MSK}& \multicolumn{1}{c}{PH2}& \multicolumn{1}{c}{SON} & \multicolumn{1}{c}{UDA} & \multicolumn{1}{c}{Average} \\ \hline
DeepAll & 0.2492 $ \pm 0.0127$  & 0.5680$\pm 0.0181$    &   0.6674$ \pm 0.0083 $   &  0.8000$ \pm 0.0167 $   &   0.8613$\pm 0.0296  $  &   0.6264$\pm 0.0312 $  &  0.6287   \\
MASF \citep{dou2019domain}   &   0.2692$ \pm 0.0146  $   &   0.5678$ \pm 0.0361$   &   0.6815$ \pm 0.0122 $   &   0.7833$\pm 0.0101 $  &   0.9204$ \pm 0.0227 $  &  0.6538$ \pm 0.0196 $   &  0.6460   \\

LDDG  \citep{li2020domain}   &   0.2793$ \pm 0.0244 $   &   0.6007$ \pm 0.0187 $  &    0.6967$\pm 0.0211  $ &  0.8167$ \pm 0.0209 $   &   0.9272$ \pm 0.0117 $  &   0.6978$ \pm 0.0182 $  &  0.6697   \\
 KDDG \citep{wang2021embracing}   &   0.3189$ \pm 0.0256 $   &   0.5829$ \pm 0.0212 $  &    0.7014$\pm 0.0178  $ &  0.9021$ \pm 0.0314 $   &   0.9398$ \pm 0.0213 $  &   0.6882$ \pm 0.0139 $  &  0.6889   \\
SWAD   \citep{cha2021swad} &  0.3582 $  \pm 0.0234$    &  0.5491 $\pm 0.0231$    &   0.6842 $\pm 0.0156$   & 0.9167 $\pm 0.0121$     &   0.9824 $\pm 0.0012$   &  0.7240 $\pm 0.0251$    &  0.7024   \\
BDIL \citep{xiao2021bit}   &   0.2985$ \pm 0.0452  $   &  \textbf{0.6204}$ \pm 0.0212 $    &   0.7059$ \pm 0.0145 $    &  0.8967$ \pm 0.0096 $   &  \underline{0.9860}$ \pm 0.0198 $    &  0.7219$ \pm 0.0284$    &  0.7049   \\
DNA \citep{chu2022dna}    &   0.3532 $\pm 0.0133$    &  0.5581 $\pm 0.0178$    &  \underline{0.7120} $\pm 0.0194$    & \underline{0.9333}$\pm 0.0045$     &  0.9851 $\pm 0.0032$    & 0.7314 $\pm 0.0141$    &  0.7122   \\
DSU  \citep{li2022uncertainty}   &    \textbf{0.3830} $\pm 0.0267$  &  0.5739 $\pm 0.0147$    &  0.6935 $\pm 0.0165$    &  0.8833 $\pm 0.0231$    &  0.9841 $\pm 0.0098$    &   0.7201 $\pm 0.0121$   &  0.7063   \\ 
MIRO \citep{cha2022domain}    &  0.3432 $ \pm 0.0092 $     & 0.5863 $ \pm 0.0113  $   &  0.6919 $ \pm 0.0101$   &   0.9300$ \pm 0.0021 $  &  0.9659 $\pm 0.0292  $  &  \underline{0.7328}  $ \pm 0.0233 $  &  0.7084 \\
Ours (in this paper)   & \underline{0.3781}$ \pm 0.0136 $     &  \underline{0.6120}$ \pm 0.0115  $   &  \textbf{0.7276} $ \pm 0.0201$   &   \textbf{0.9416}$ \pm 0.0103 $  &   \textbf{0.9889}$\pm 0.0041  $  &   \textbf{0.7486} $ \pm 0.0123 $  &  \textbf{0.7328} \\\bottomrule[1.2pt] 
\end{tabular}
\end{adjustbox}
\label{skin_exp}
\end{table*}

\textbf{Discussion. } The rationale that our proposed method can benefit  DG performance on small-data scenario can come from two aspects. First, \textit{BNN can be adaptive to insufficient data well compared with deterministic models} \citep{graves2011practical,mallick2021deep}. For our proposed method, BNN is introduced to the DG problem in the context of insufficient data, where BNN-based feature extractor and classification layers can take both consistent improvements (see 2nd and 3rd columns in Table \ref{ablation}). More importantly, \textit{domain-invariant representation learning under this probabilistic framework from global and local perspectives} contributes to more robust cross-domain performance (see 4th and 5th columns in \hyperref[ablation]{Table 4}; \hyperref[parameter_modification]{Figure 3} in \hyperref[kme]{Section 4.5} for the effectiveness of P-MMD).

\section{Experiments}
We evaluate our proposed method on three medical imaging tasks: 1) epithelium stroma classification, 2) skin lesion classification, 3) spinal cord gray matter segmentation. The used datasets in these tasks are collected from different healthcare institutes and suffer from the domain shift problem in the context of insufficient samples, \textit{i.e. insufficient sample scenarios exist either in all or some source
domains.}

\subsection{Epithelium Stroma Classification}
Epithelium stroma classification is a fundamental step for the prognostic analysis of the tumor. The public histopathological image datasets for binary classification (epithelium or stroma) are collected from three healthcare centers with different staining types and tissue densities\footnote{http://fimm.webmicroscope.net/supplements/epistroma}: IHC, NKI, and VGH. After the patching operation, IHC, NKI and VGH datasets respectively have 1342,1230, and 1376 patches, which means that the insufficient sample problem exist in \textit{all source domains} compared with large-scale natural images. We randomly split the data of each source domain into a training set (80\%) and a test set (20\%) and adopt the leave-one-domain-out strategy for evaluation. The pretrained ResNet18 is introduced as the backbone. The structure of Bayesian layer in $Q_{\phi}$ is a fully-connected-based BNN with $512 \times 512$. The structure of Bayesian layer in $C_{\omega}$ is also a fully-connected-based BNN with $512 \times 2$. We utilize Adam optimizer with learning rate as $5 \times 10^{-5}$ for training. The batch size is 32 for each source domain with 4000 iterations. The hyperparameters are selected in a wide range on the validation set, where the $\beta_{1}$ and $\beta_{2}$ are 0.1 and 0.7 for the $\mathcal{L}_{local}$ and the $\mathcal{L}_{global}$, respectively. For the P-MMD, \textit{level-1} and \textit{level-2} kernels are the Gaussian RBF kernels (the kernel bandwidth is empirically set to $1$ for all kernels) by following \citet{muandet2012learning}. For the P-CSA loss, the distance margin $\xi$ is set to 1. By balancing the performance and computational efficiency, the number of MC sampling in each Bayesian layer (\textit{a.k.a.}, $T$), is set to 10. We also discuss the influence of using different $T$ in the section \ref{mceffective}. We report the results based on average value and standard deviation in each target domain by running the experiment for five different times.

\begin{table*}[!h]
\renewcommand{\arraystretch}{1.1}
  \centering
  \caption{Domain generalization results on gray matter segmentation task. For the DSC, CC, TPR, and JI, the higher the better. For the ASD, the lower the better.  Note that all baseline methods adopt the SWAD method \citep{cha2021swad} for weight averaging. The baseline, namely SWAD, denotes the ERM training strategy with the SWAD method.}
    \scalebox{0.7}{
    \subtable[MASF]{
    \begin{tabular}{cc|ccccc}
    \toprule
    source & target & DSC   & CC    & JI & TPR   & ASD \\
    \midrule
    2,3,4 & 1     & 0.8502 & 64.22 & 0.7415 & 0.8903 & 0.2274 \\
    1,3,4 & 2     & 0.8115 & 53.04 & 0.6844 & 0.8161 & 0.0826 \\
    1,2,4 & 3     & 0.5285 & -99.3 & 0.3665 & 0.5155 & 1.8554 \\
    1,2,3 & 4     & \textbf{0.8938} & \textbf{76.14} & \textbf{0.8083} & \underline{0.8991} & 0.0366  \\
    \midrule
    \multicolumn{2}{c|}{Average} & 0.7710 & 23.52 & 0.6502 & 0.7803 & 0.5505  \\
    \bottomrule
    \end{tabular}%
    }

    
    \subtable[KDDG]{
    \begin{tabular}{cc|ccccc}
    \toprule
    source & target & DSC   & CC    & JI & TPR   & ASD \\
    \midrule
     2,3,4 & 1     & \underline{0.8745} & \underline{70.75} & \underline{0.7795} & 0.8949 & 0.0539 \\
     1,3,4 & 2     & 0.8229 & 56.71 & 0.6997 & 0.8226 & 0.0490 \\
     1,2,4 & 3     & \bf{0.5676} & \bf{-63.1} & 0.3866 & 0.5904 & \underline{1.2805} \\
     1,2,3 & 4     & 0.8894 & 75.06 & 0.8011 & 0.9222 & 0.0377  \\
    \midrule
    \multicolumn{2}{c|}{Average} & 0.7886 & 34.86 & 0.6667 & 0.8075 & 0.3553  \\
    \bottomrule
    \end{tabular}%
    }

  }  
 
  \scalebox{0.7}{
    \subtable[LDDG]{
    \begin{tabular}{cc|ccccc}
    \toprule
    source & target & DSC   & CC    & JI & TPR   & ASD \\
    \midrule
    2,3,4 & 1     & 0.8708 & 69.29 & 0.7753 & 0.8978 & \textbf{0.0411}\\
    1,3,4 & 2     & 0.8364 & 60.58 & 0.7199 & \underline{0.8485} & 0.0416 \\
    1,2,4 & 3     & 0.5543 & -71.6 & \underline{0.3889} & \underline{0.5923} & 1.5187  \\
    1,2,3 & 4     & 0.8910 & 75.46 & 0.8039 & 0.8844 & \textbf{0.0289} \\
    \midrule
    \multicolumn{2}{c|}{Average} & 0.7881 & 33.43 & 0.6720 & 0.8058 & 0.4076  \\
    \bottomrule
    \end{tabular}%
    }

    
    \subtable[\textcolor{black}{SWAD}]{
    \begin{tabular}{cc|ccccc}
    \toprule
    source & target & DSC   & CC    & JI & TPR   & ASD \\
    \midrule
    2,3,4 & 1     & 0.8726 & 70.23 & 0.7702 & 0.8995 & 0.0502\\
    1,3,4 & 2     & 0.8378 & 60.71 & 0.7230 & 0.8176 & 0.0424 \\
    1,2,4 & 3     & 0.5388 & -99.0 & 0.3789 & 0.5083 & 1.4789  \\
    1,2,3 & 4     & 0.8903 & \underline{75.89} & 0.8026 & 0.8859 & \underline{0.0302} \\
    \midrule
    \multicolumn{2}{c|}{Average} & 0.7849 & 26.96 & 0.6687 & 0.7778 & 0.4002  \\
    \bottomrule
    \end{tabular}%
    }
  }  
  \scalebox{0.7}{


    \subtable[DSU]{
    \begin{tabular}{cc|ccccc}
    \toprule
    source & target & DSC   & CC    & JI & TPR   & ASD \\
    \midrule
    2,3,4 & 1     & 0.8739 & 70.32 & 0.7794 & \underline{0.9210} & 0.0793 \\
    1,3,4 & 2     & 0.8474 & 63.58 & 0.7367 & \textbf{0.8502} & 0.0494 \\
    1,2,4 & 3     & 0.5574 & -70.4 & 0.3923 & 0.6097 & 1.5049 \\
    1,2,3 & 4     & 0.8897 & 75.10 & 0.8018 &  0.9225 & 0.0415  \\
    \midrule
    \multicolumn{2}{c|}{Average} & 0.7921 & 34.65 & 0.6775 & 0.8225 & 0.4362  \\
    \bottomrule
    \end{tabular}%
    }
    \subtable[Ours]{
    \begin{tabular}{cc|ccccc}
    \toprule
    source & target & DSC   & CC    & JI & TPR   & ASD \\
    \midrule
    2,3,4 & 1     & \textbf{0.8786} & \textbf{71.57} & \textbf{0.7873} & \textbf{0.9293} & \underline{0.0422} \\
    1,3,4 & 2     & \textbf{0.8485} & \textbf{63.78}  & \textbf{0.7389} &  0.8401 & \textbf{0.0401} \\
    1,2,4 & 3     &  \underline{0.5634} & \underline{-68.0} & \textbf{0.3992}  & \textbf{0.6103}  & \textbf{1.2239} \\
    1,2,3 & 4     &  \underline{0.8921}& 75.69 & \underline{0.8058} & \textbf{0.9245} &  0.0362  \\
    \midrule
    \multicolumn{2}{c|}{Average} & \textbf{0.7957} & \textbf{35.76} & \textbf{0.6828} & \textbf{0.8260} &  \textbf{0.3356} \\
    \bottomrule
    \end{tabular}%
    }
  }

\label{gm_full_results}%
\end{table*}
\textbf{Results.} Table \ref{breast_exp} shows the  epithelium stroma classification results on different target domains. We compare with five different approaches. All methods are constructed via a SWAD-based  \citep{cha2021swad} framework (where a pretrained ResNet18 as the backbone) with the same training schedule. DeepAll is a deterministic version of our proposed method that is trained on all source domains without any DG strategy in the sequel. Some observations can be summarized as following.  
First, both our proposed method and BDIL \citep{xiao2021bit} achieve promising results on both IHC and VGH tasks, which may benefit from the positive impact of probabilistic framework on insufficient samples.
However, BDIL has an obvious performance drop on the more challenging NKI domain (which has a lower average accuracy compared with other target domains) compared with our proposed method (\textit{i.e.}, 71.89\% \textit{v.s.} \textbf{76.71}\%). This may be due to  the introduction of global alignment (\textit{i.e.}, P-MMD, as used by ours), 
which is more powerful for learning domain-invariant representations than only using local negative pairs (as used by BDIL). Second, compared with contrastive semantic alignment-based method (\textit{i.e.}, MSFA \citep{dou2019domain}), our proposed method achieves a significantly better performance (80.45\% \textit{v.s.} \textbf{88.82}\% on IHC domain), due to a more reliable distribution-based contrastive learning manner on insufficient samples from all source domains. Finally, our proposed method achieves the best average performance with a clear margin compared with other approaches (\textit{i.e.}, LDDG \citep{li2020domain}, KDDG \citep{wang2021embracing}, DNA \citep{chu2022dna}, MIRO \citep{cha2022domain}, and DSU \citep{chu2022dna}).

\subsection{Skin Lesion Classification}
 Seven public skin lesion datasets\footnote{https://challenge.isic-archive.com/landing/2018/47/} for seven classes of lesions are collected from various institutes using different dermatoscope types: HAM10000 with 10015 images, 
UDA with 601 images, SON with 9251 images, DMF with 1212 images, MSK with 3551 images, D7P with 1926 images, and PH2 with 200 images. We can observe that the insufficient sample problem exists in \textit{some source domains}, especially in PH2 and UDA domains. Following previous work \citep{li2020domain}, each domain is randomly split into a 50\% training set, 30\% test set, and 20\% validate set, respectively. As adopted in \citet{li2020domain}, one domain from DMF, D7P, MSK, PH2, SON and UDA is as target domain and the
remaining domains together with HAM10000 as source domains. The pretrained ResNet18 is introduced as the backbone. The structure of Bayesian layer in $Q_{\phi}$ is a fully-connected-based BNN with $512 \times 512$. The structure of Bayesian layer in $C_{\omega}$ is also a fully-connected-based BNN with $512 \times 7$. The Adam optimizer is employed with learning rate of $5 \times 10^{-5}$ for 2000 iterations. The hyperparameters are selected in a wide range on the validation set, where the $\beta_{1}$ and $\beta_{2}$ are 0.1 and 0.7 for the $\mathcal{L}_{local}$ and the $\mathcal{L}_{global}$, respectively. The batch size is 32 for each source domain. The remaining settings are the same with epithelium stroma classification. For the results, the average value and standard deviation  are reported by running five times.
\begin{table*}
\centering
\caption{Ablation study on each component of our proposed method for spinal cord gray matter segmentation task (where ``site2" is as the target domain). The model on the first row denotes the basic Unet model. }
    \begin{adjustbox}{width=0.75\textwidth}
		\begin{tabular}{lclccccccc}
\hline
\multicolumn{1}{c}{\begin{tabular}[c]{@{}c@{}}Backbone\\ (Unet)\end{tabular}} &
  \begin{tabular}[c]{@{}c@{}}Bayesian \\ Layers\end{tabular} &
  \begin{tabular}[c]{@{}c@{}}Local \\ Alignment\end{tabular}
    &\begin{tabular}[c]{@{}c@{}}Global \\ Alignment\end{tabular} &

  \begin{tabular}[c]{@{}l@{}}Bayesian\\ Classifier\end{tabular} &
  DSC &
  CC &
  JI &
  TPR &
  ASD \\ \hline
\multicolumn{1}{c}{\CheckmarkBold} & \multicolumn{1}{c|}{\XSolidBrush}     &   \multicolumn{1}{c}{-}                   &  \multicolumn{1}{c|}{-}        & \multicolumn{1}{c}{\XSolidBrush}  & 0.7223          & 26.21          & 0.5789          & 0.8109          & 0.0992          \\ \multicolumn{1}{c}{\CheckmarkBold}
 & \multicolumn{1}{c|}{\CheckmarkBold}
    &
  \multicolumn{1}{c}{-} &
  \multicolumn{1}{c|}{-} &
  \multicolumn{1}{c}{\XSolidBrush} &
  \multicolumn{1}{c}{0.7934} &
  \multicolumn{1}{c}{47.19} &
  \multicolumn{1}{c}{0.6595} &
  \multicolumn{1}{l}{0.8133} &
  \multicolumn{1}{c}{0.0692} \\ \multicolumn{1}{c}{\CheckmarkBold}
 & \multicolumn{1}{c|}{\CheckmarkBold}    & \multicolumn{1}{c}{-} &       \multicolumn{1}{c|}{-}   & \multicolumn{1}{c}{\CheckmarkBold}  & 0.8268          & 57.52          & 0.7067          & 0.8156          & 0.0501          \\ \multicolumn{1}{c}{\CheckmarkBold}
 & \multicolumn{1}{c|}{\CheckmarkBold}  &  \multicolumn{1}{c}{\textcolor{red}{\CheckmarkBold}                      }& \multicolumn{1}{c|}{\XSolidBrush}& \multicolumn{1}{c}{\CheckmarkBold}  & 0.8364          & \underline{60.72}          & 0.7195          & \underline{0.8267}          & 0.0486          \\ \multicolumn{1}{c}{\CheckmarkBold}
 & \multicolumn{1}{c|}{\CheckmarkBold}            &   \multicolumn{1}{c}{\XSolidBrush}        & \multicolumn{1}{c|}{\textcolor{red}{\CheckmarkBold}                      }  &   \multicolumn{1}{c}{\CheckmarkBold}               &     \underline{0.8371}           &   60.57             &     \underline{0.7217}            &            0.8152    & 0.0510 \\ \multicolumn{1}{c}{\CheckmarkBold} 
 & \multicolumn{1}{c|}{\CheckmarkBold}     &   \multicolumn{1}{c}{\textcolor{red}{\CheckmarkBold}                      }                    &\multicolumn{1}{c|}{\textcolor{red}{\CheckmarkBold}                      }     & \multicolumn{1}{c}{\CheckmarkBold} & \textbf{0.8485} & \textbf{63.78} & \textbf{0.7389} & \textbf{0.8401} & \textbf{0.0401} \\ \hline
\end{tabular}
        \end{adjustbox}
        \label{ablation}
\end{table*}

\textbf{Results.} Table \ref{skin_exp} shows the skin lesion classification accuracies on different target domains. Six different methods are utilized for comparison. All approaches are implemented using the SWAD-based framework and a pretrained ResNet18 model as the backbone. One has some observations as following. First, compared with contrastive semantic alignment-based method (\textit{e.g.}, MASF), our proposed method provides more reliable distribution-based pairs, leading to a better performance on 
insufficient samples from \textit{some source domains} (MASF:0.2692 \textit{v.s.} Ours: \textbf{0.3781} on DMF domain, where PH2 and UDA as the parts of source domain). Second, although BDIL slightly outperforms our proposed method on D7P domain (\textbf{0.6204} \textit{v.s.} 0.6120), our proposed method has a significantly better performance on the challenging DMF domain (0.2985 \textit{v.s.} \textbf{0.3781}) and the average results (0.7049 \textit{v.s.} \textbf{0.7328}) and other domains. Third, it seems that other baseline methods (\textit{e.g.}, SWAD, DNS, and DSU) impose respective schemes to relieve the impact of insufficient samples from \textit{some source domains}. For example, DSU achieves the best performance on DMF domain. This may be due to the positive impact of straightforward domain randomization. Yet, it is  difficult for DSU to realize consistently better results in multiple domains compared with our proposed method,  owing to the lack of explicit domain alignment. 

\begin{figure*}[!h]
    \centering
    \includegraphics[width=0.75\textwidth]{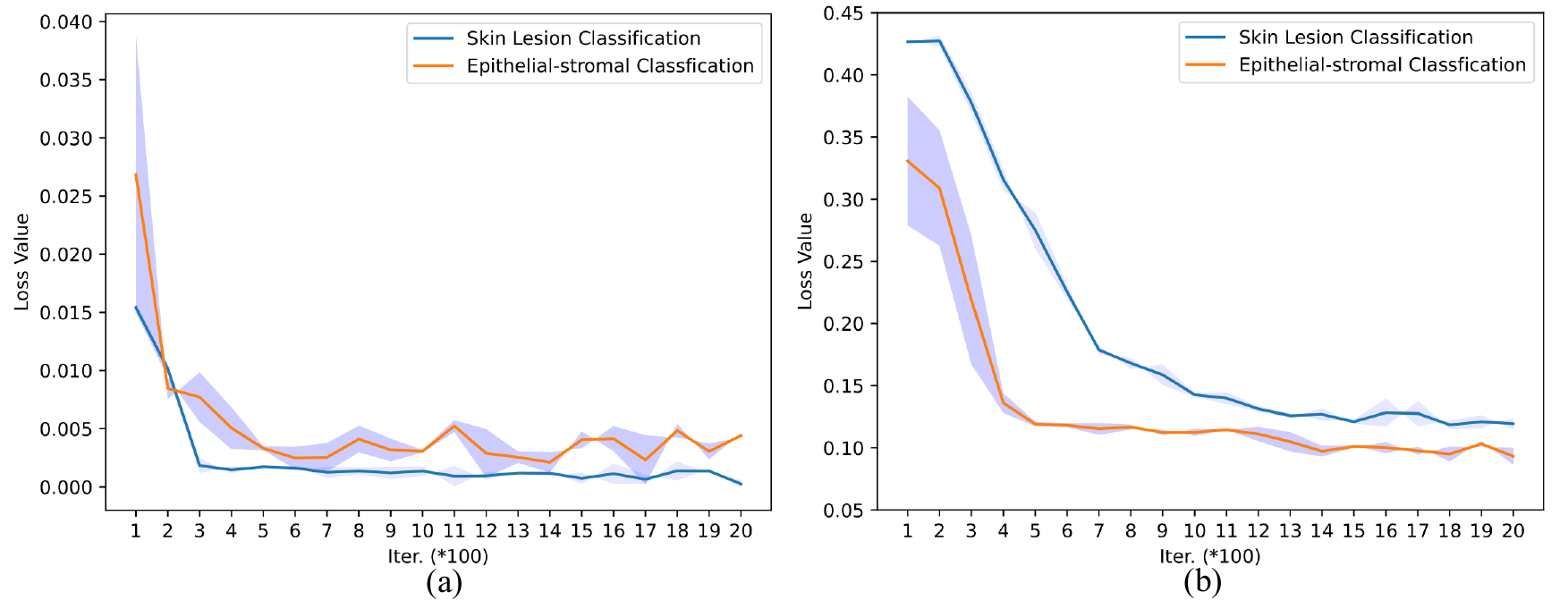}
    \caption{The loss curve of iteration on skin lesion and epothelial-stromal classficaiton tasks. (a) Global alignment loss (b) Local alignment loss. }
    \label{loss}
\end{figure*}

\subsection{Spinal Cord Gray Matter Segmentation} \label{scgms_experiment}
The spinal cord gray matter (GM) segmentation (pixel-level classification) is an emergent task for predicting disability via evaluating the atrophy of GM area. The acquired magnetic resonance imaging (MRI)
data are collected from four healthcare centers\footnote{http://niftyweb.cs.ucl.ac.uk/challenge/index.php}:``site1" with 30 slices, ``site2" with 113 slices, ``site3" with 246 slices, and ``site4" with 122 slices.  One can observe that the insufficient sample problem exists in \textit{some source domains}, especially in ``site1" domain. By following previous work \citep{li2020domain}, we randomly split the data of each source domain into a training set (80\%) and a test set (20\%) and adopt the leave-one-domain-out strategy for evaluation. The hyperparameters are selected in a wide range on the validation set, where the $\beta_{1}$ and $\beta_{2}$ are 0.01 and 0.001 for the $\mathcal{L}_{local}$ and the $\mathcal{L}_{global}$, respectively. The 2D-Unet \citep{ronneberger2015u} is leveraged as the backbone for all methods. The structures of $Q_{\phi}$ and  $C_{\omega}$ as well as more experimental details can be found in  \hyperref[details]{Appendix B}. By following \citet{li2020domain}, the average results in each target domain are reported by running three times.

\begin{table*}[!t]
\centering
\renewcommand{\arraystretch}{1.4}
\caption{Domain generalization results on MSK dataset by randomly picking same proportion of samples from each source domain. A smaller proportion ($< 40\%$) is unavailable because equal batch sizes cannot be maintained in PH2 dataset.}
\begin{adjustbox}{max width=0.65\textwidth}
\begin{tabular}{cccc}
\toprule[1pt]
\textbf{ Proportion (\%)} &
  BDIL &
  DNA &
  Ours \\ \hline
\textbf{100} & 0.7059$\pm$ 0.0284 & 0.7121$\pm$ 0.0141  & \textbf{0.7276}$\pm$0.0123  \\ \hline \hline
\textbf{80}  & \multicolumn{1}{l}{0.6625$\pm$0.0920}   & 0.6591$\pm$0.0022 & \textbf{0.6975}$\pm$0.0036  \\
\textbf{60}  & 0.6468$\pm$0.0106 & 0.6149$\pm$0.0112 & \textbf{0.6641}$\pm$0.0114  \\
\textbf{40}  & 0.6491$\pm$0.0171 & 0.6065$\pm$0.0111 & \textbf{0.6579}$\pm$0 .0057 \\ \hline
\textbf{Average (80,60,40) $\uparrow$ }                        & 0.6528             & 0.6268             & \textbf{0.6732}                               \\ \hline
\textbf{Average Attenuation Rate $\downarrow$}        & 7.67\%            & 11.98\%           & \textbf{7.37\%} \\ \bottomrule[1pt]                        
\end{tabular}
\end{adjustbox}
\label{challenging}

\end{table*}

\begin{table*}[!t]
\centering
\renewcommand{\arraystretch}{1.4}
\caption{Domain generalization results on MSK dataset by randomly picking same number of samples from each class in each domain.}
\begin{adjustbox}{max width=0.65\textwidth}
\begin{tabular}{cccc}
\toprule[1pt]
\textbf{ Number of sample} &
  BDIL &
  DNA &
  Ours \\ \hline
\textbf{40}   & 0.5897 $\pm$ 0.0029 & 0.5412 $\pm$ 0.0143 & \textbf{0.6368} $\pm$ 0.0074 \\
\textbf{30}  & 0.5762 $\pm$ 0.0101 & 0.5132 $\pm$ 0.0229 & \textbf{0.6138} $\pm$ 0.0291 \\
\textbf{20} & 0.5573 $\pm$ 0.0011 & 0.5048 $\pm$ 0.0087 & \textbf{0.6037} $\pm$ 0.0121 \\ \hline
\textbf{Average (40,30,20) $\uparrow$ }                        & 0.5744             & 0.5196             & \textbf{0.6183}                               \\ \hline
\textbf{Average Attenuation Rate $\downarrow$}        & 5.49\%            & 6.72\%           & \textbf{5.19\%} \\ \bottomrule[1pt]                        
\end{tabular}
\end{adjustbox}
\label{Fixed size of samples per class}

\end{table*}

\textbf{Results.} Table \ref{gm_full_results} shows the spinal cord GM segmentation results on different target domains. Dice Similarity Coefficient (DSC), Jaccard Index (JI), and Conformity Coefficient (CC) are used to measure the accuracy of obtained segmentation results. True Positive Rate (TPR) and Average Surface Distance (ASD) are introduced from statistical and distance-based perspectives. We compare with five different methods (which have effective segmentation performance). Some observations can be found as follows. First, our proposed method outperforms all baseline methods in terms of average results for five quantitative metrics. Second, compared with the contrastive semantic alignment-based method (\textit{e.g.}, MASF), reliable pixel-level pairs constructed by our proposed method also contribute to improving performance in scenarios with insufficient samples from some source domains. Third, as we can see,  our proposed method and DSU achieve the best and second-best performance compared with other baselines, which may be reasonable as they can benefit from the modeling of the data uncertainty in the small-data regime using the BNN or the distribution modeling.

\subsection{Ablation Analysis.}  The spinal cord gray matter segmentation task is utilized to explore the effectiveness of each component for our proposed method, \textit{due to its various quantitative metrics}. The results can be shown in Table \ref{ablation}. First, we observe that better performance can be achieved by introducing a probabilistic layer compared with the results that using Unet, which reflects the superiority of probabilistic models. Secondly, we observe that by either introducing local or global alignment for domain-invariant information learning, better performance can be achieved compared with the results of only using the probabilistic layer, which shows the effectiveness of the introduced probabilistic feature regularization term. Last but not least, by imposing domain-invariant learning with both local and global views, the performances are further improved, which justifies the effectiveness of our proposed method by jointly considering local and global alignment.

\textbf{Effectiveness of domain-invariant loss.} We are also interested in the impacts of domain-invariant losses on different tasks. Thus, we visualize the learning curves of different task in Figure \ref{loss}. As we can observe,  for the skin lesion (on DMF) and epithelium-stroma (on IHC) classification tasks, the loss curves with iterations reflect the global discrepancy converges faster than the local discrepancy, while the more challenging cross-domain task converges more slowly on global alignment.
\begin{figure*}[!t]
  \centering
  \subfigure[Local alignment]{

  \begin{minipage}{0.45\textwidth}
  \centering
     \centerline{\includegraphics[width= \textwidth, height=4.5cm]{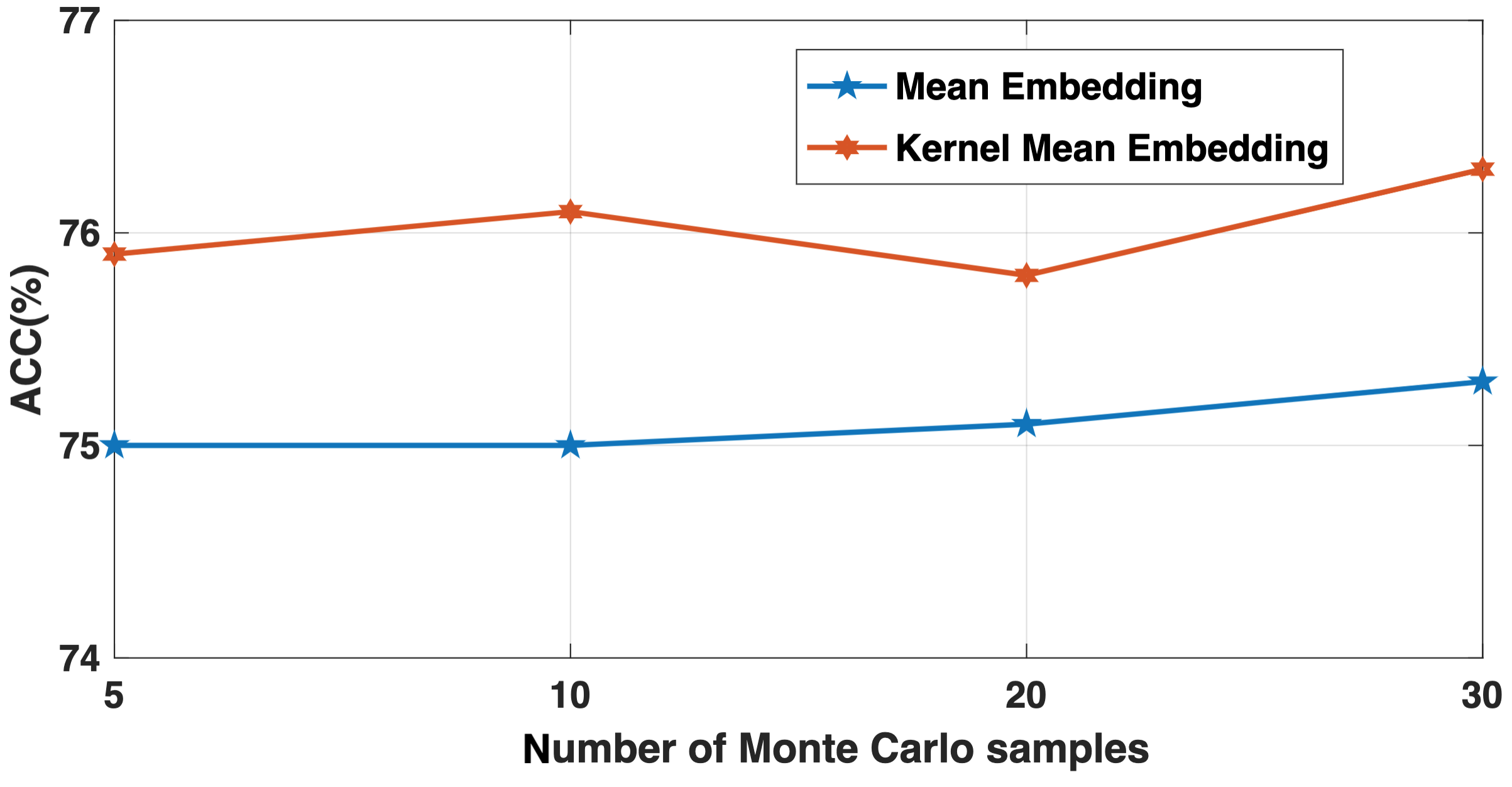}}
  \end{minipage}
  }
  \subfigure[Global alignment]{

  \begin{minipage}{0.45\textwidth}
  \centering

     \centerline{\includegraphics[width= \textwidth,height=4.5cm]{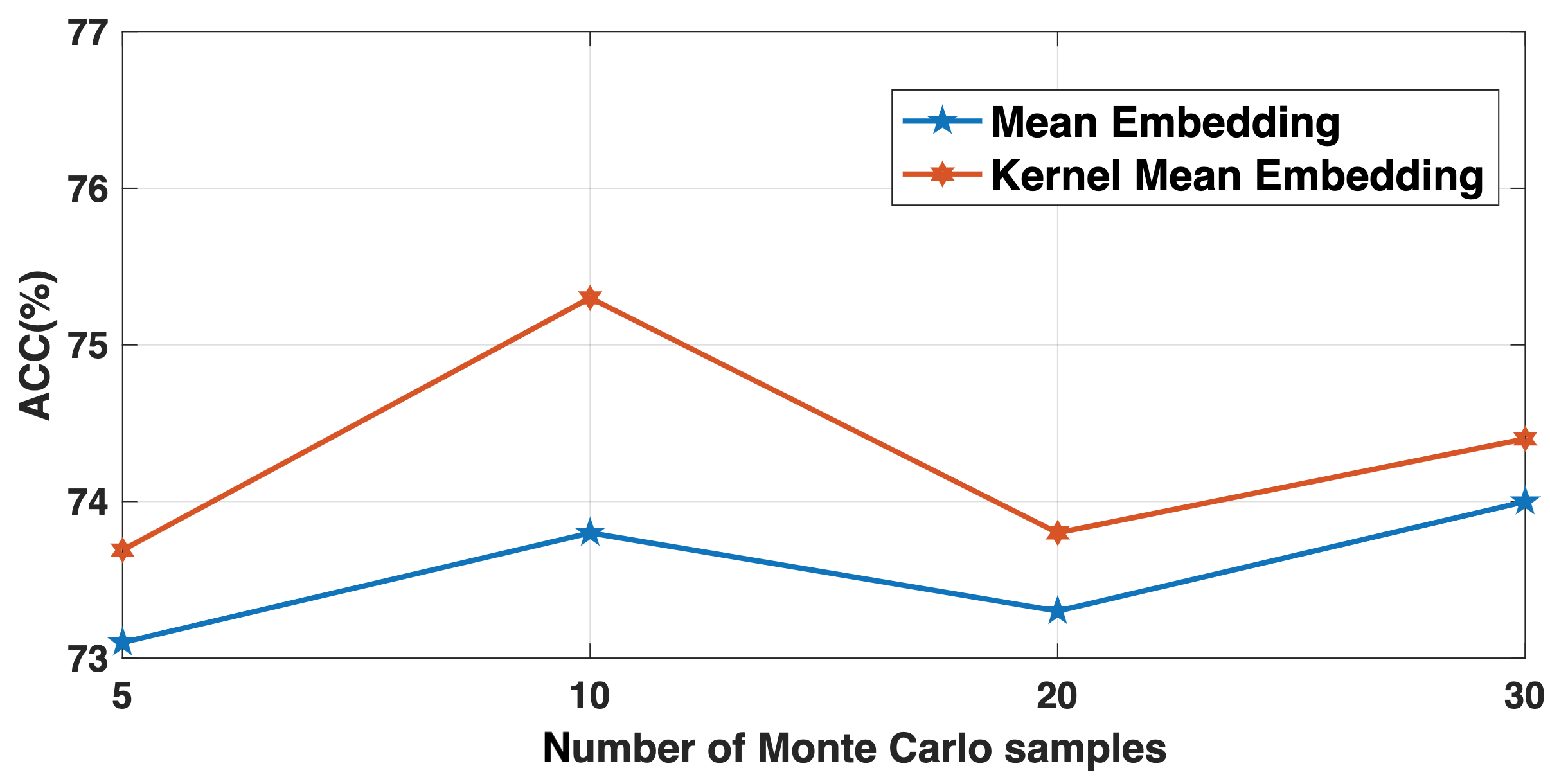}}
  \end{minipage}
  }
  \caption{The performance comparison between mean embedding method and kernel mean embedding method with different Monte Carlo samples $T$. For each sub-figure, we use only one alignment operation. (a) Local alignment. \textbf{Mean Embedding:} The mean embedding operation with Euclidean distance is utilized between probabilistic embedding pairs. \textbf{Kernel Mean Embedding:} The kernel mean embedding with MMD distance is utilized between probabilistic embedding pairs. (b) Global alignment. \textbf{Mean Embedding:} The mean embedding operation with MMD distance is utilized between domains (as distributions). \textbf{Kernel Mean Embedding:} The kernel mean embedding with P-MMD distance is utilized between domains (as distributions over distributions).}
  \label{parameter_modification}
\end{figure*}
\subsection{Further analyses of our proposed method}

\textbf{Results on Different Fractions of Training Data.}We are interested in how different fractions of training samples influence the final performance based on small-data scenario. To this end, we adopt skin lesion classification task for evaluation since the number of samples in each domain turns out to be imbalance (some domains such as HAM10000 contain sufficient data while the number of samples in some other domains such as PH2 and UDA is insufficient). As such, we can better simulate the scenario that the issue of insufficient samples either exists in \textit{all source domains} or in \textit{partial source domains}. We choose MSK dataset with the second-highest number of samples as the target domain and the remaining datasets as the source domains (including PH2 and UDA). Table \ref{challenging} shows the accuracies. As we can see, compared with DNA (with second-best performance on 100\% of samples), our proposed method and BDIL have better average results and lower performance attenuation as the decrease of sample number, due to their probabilistic gain under small data scenarios. Compared with BDIL, our proposed method shows a more robust performance on this challenging small data scenario, due to the integration of local (P-CSA) and global (P-MMD) alignments.

\textbf{Results on Different Numbers of Samples for Training Data.} We are also interested in how different   numbers of samples per class influence the final performance based on small-data scenarios. To this end, we also adopt skin lesion classification task for evaluation. Specifically, we randomly draw T samples from each class in a source domain to represent this domain for training. Here, we set T to 20,30, and 40, respectively, in different experiments.

The results can be found in Table \ref{Fixed size of samples per class}. As we can see,  our proposed method achieved the best performance among all settings compared with all baseline methods. Meanwhile,  it seems that the Bayesian-based DG approaches (e.g., our proposed method and BIDL) have better performance compared with other methods, which is reasonable as the BNN can be adaptive to the small data scenario well. Especially, our proposed method has around 5\% improvements compared with the second-best method when $T$ is set to smaller, i.e., 20.

\textbf{Kernel Mean Embedding (\textit{i.e.}, P-MMD) \textit{v.s.} \textbf{Mean Embedding}.\label{kme}}
We explore the effect of different schemes for probabilistic embeddings. A straightforward method is to represent probabilistic embeddings with the expectation, which is called the ``Mean Embedding".  Then, a probabilistic embedding can be considered as a latent point and the MMD can be used to measure the discrepancy between distributions consisting of latent points.
\begin{table*}[!t]
\renewcommand{\arraystretch}{1.1}
\small
\centering
\caption{\textbf{Out-of-domain accuracies (\%) on} PACS based on ResNet50.} 
\label{table:pacs}
\begin{tabular}{ccccc|c}
\toprule[1pt]
\multicolumn{1}{c}{Method}  & Art & Cartoon & Photo & Sketch & Average (\%)  \\ \hline

RSC \citep{huang2020self}  & 78.9 &  76.9 & 94.1 & 76.8 & 81.7 \\
L2A-OT \citep{zhou2020learning}   & 83.3 & 78.2 & 96.2 & 73.6 & 82.8 \\
MatchDG \citep{mahajan2021domain}  & 81.2 & 80.4 & 96.8 & 77.2 & 83.9 \\
pAdaIN \citep{nuriel2021permuted}  & 81.7 & 76.6 & 96.3 & 75.1 & 82.5 \\
MixStyle \citep{zhou2021domain}  & 86.8 & 79.0 & 96.6 & 78.5 & 85.2 \\

SagNet \citep{nam2021reducing}& 87.4 & 80.7 & 97.1 & 80.0 & 86.3 \\
ERM \citep{vapnik1999overview} & 84.7& 80.8 & 97.2 & 79.3 & 85.5 \\
DNA \citep{chu2022dna} & 89.8 & 83.4 & 97.7 & 82.6 & 88.4 
 \\ \hline

ERM+SWAD \citep{cha2021swad} & 89.3 & 83.4 & 97.3 & 82.5 & 88.1 \\
MIRO+SWAD \citep{cha2022domain} & - & - & - & - & 88.4 \\
\textbf{Ours+SWAD}& \textbf{90.2} & \textbf{85.2}  & \textbf{98.7} & \textbf{83.6} & \textbf{89.4}\\
\bottomrule[1pt]
\end{tabular}

\label{pacs}
\end{table*}

\begin{table*}[!t]
\centering
\small
\renewcommand{\arraystretch}{1.1}
\caption{\textbf{Out-of-domain accuracies (\%) on} OfficeHome based on ResNet50.}
\begin{tabular}{ccccc|c}
\toprule
\textbf{Algorithm} & \textbf{Art} & \textbf{Clipart} & \textbf{Product} & \textbf{Real} & \textbf{Avg} \\
\midrule
Mixstyle \citep{zhou2021domain} &  51.1  & 53.2  & 68.2  & 69.2  & 60.4 \\
RSC \citep{huang2020self} & 60.7  & 51.4  & 74.8  & 75.1  & 65.5 \\
DANN \citep{ganin2016domain} & 59.9  & 53.0  & 73.6  & 76.9  & 65.9 \\
GroupDRO \citep{sagawa2019distributionally}& 60.4  & 52.7  & 75.0  & 76.0  & 66.0 \\
MTL \citep{blanchard2021domain} & 61.5  & 52.4  & 74.9  & 76.8  & 66.4 \\
VREx \citep{krueger2021out} & 60.7  & 53.0  & 75.3  & 76.6  & 66.4 \\
MLDG \citep{balaji2018metareg} & 61.5  & 53.2  & 75.0  & 77.5  & 66.8 \\
SagNet \citep{qian2021latent} &63.4  & 54.8  & 75.8  & 78.3  & 68.1 \\
CORAL \citep{sun2016deep} &65.3  & 54.4  & 76.5  & 78.4  & 68.7 \\
ERM \citep{vapnik1999overview} & 63.1  &51.9  & 77.2  & 78.1  & 67.6\\
DNA \citep{chu2022dna} &  67.7 &57.7  & 78.9  & 80.5  & 71.2
\\

 \hline
 ERM+SWAD \citep{cha2021swad} & 66.1  &57.7  & 78.4  & 80.2  & 70.6\\
 MIRO+SWAD \citep{cha2022domain} & - & - & - & - & \textbf{72.4} \\
  \textbf{Ours+SWAD} &   68.2 & 58.9 & 80.2 & 80.7 & \underline{72.0}\\
\bottomrule
\end{tabular}
\label{officehome}
\end{table*}

\begin{table*}[!t]
\centering
\small
\renewcommand{\arraystretch}{1.1}
\caption{\textbf{Out-of-domain accuracies (\%) on} VLCS based on ResNet50.}
\begin{tabular}{ccccc|c}
\toprule
\textbf{Algorithm} & \textbf{Caltech} & \textbf{LabelMe} & \textbf{SUN} & \textbf{VOC} & \textbf{Avg} \\
\midrule
Mixstyle \citep{zhou2021domain}& 98.3 & 64.8 & 72.1 & 74.3 & 77.4 \\
RSC \citep{huang2020self}& 97.9 & 62.5 & 72.3 & 75.6 & 77.1 \\
DANN \citep{ganin2016domain}& 99.0 & 65.1 & 73.1 & 77.2 & 78.6 \\
GroupDRO \citep{sagawa2019distributionally}& 97.3 & 63.4 & 69.5 & 76.7 & 76.7 \\
MTL \citep{blanchard2021domain}& 97.8 & 64.3 & 71.5 & 75.3 & 77.2 \\
VREx \citep{krueger2021out}& 98.4 & 64.4 & 74.1 & 76.2 & 78.3 \\
MLDG \citep{balaji2018metareg}& 97.4 & 65.2 & 71.0 & 75.3 & 77.2 \\
SagNet \citep{qian2021latent}&97.9 & 64.5 & 71.4 & 77.5 & 77.8 \\
CORAL \citep{sun2016deep}&98.3 & 66.1& 73.4 & 77.5 & 78.8 \\
ERM \citep{vapnik1999overview}& 97.7 & 64.3 & 73.4 & 74.6 & 77.5\\
DNA \citep{chu2022dna}& 98.8 & 63.6 & 74.1 & 79.5 & 79.0
\\
\hline
ERM+SWAD \citep{cha2021swad}& 98.8 & 63.3 & 75.3 & 79.2 & 79.1\\
MIRO+SWAD \citep{cha2022domain} & - & - & - & - & \textbf{79.6} \\
\textbf{Ours+SWAD} & 98.9 & 63.4 & 75.8 & 79.8 & \underline{79.5}\\
\bottomrule
\end{tabular}
\label{VLCS}
\end{table*}
For the mean embedding-based $\mathcal{L}_{global}$, the computational process of this scheme for MMD distance can be formulated as
\begin{footnotesize}
    \begin{eqnarray}\label{m-mmd}
    \operatorname{MMD}(\mathbb{P}_{l}, \mathbb{P}_{t})^{2} & = & \|\frac{1}{n_{l}} \sum_{i=1}^{n_{l}} \varphi(\mathbb{E}[\Pi_{l_{i}}])-\frac{1}{n_{t}} \sum_{j=1}^{n_{t}} \varphi(\mathbb{E}[\Pi_{t_{j}}])\|_{\mathcal{H}}^{2}. \nonumber\\ 
\end{eqnarray}
\end{footnotesize} Eqation (\ref{m-mmd}) can be further constructed a global alignment loss $\mathcal{L}_{global}$. For the local alignment loss $\mathcal{L}_{local}$, the Euclidean distance can be used to compute the distance between latent points, which is similar to the original CAS loss in \citet{motiian2017unified}. For the positive pairs with the same label, the mean embedding-based positive contrastive loss  can be represented as 
\begin{footnotesize}
    \begin{eqnarray}
    \mathcal{L}_{local}^{pos} = \frac{1}{2} \|\frac{1}{T} \sum_{i=1}^{T} \mathbb{E}\left[M_{\Theta}(\mathbf{z}_{n_{i}})]\right)-\frac{1}{T} \sum_{j=1}^{T} \mathbb{E}\left[M_{\Theta}(\mathbf{z}_{q_{j}})]\right)\|_{2}^{2}. \nonumber \\
\end{eqnarray}
\end{footnotesize}$M_{\Theta}(\cdot)$ denotes the embedding network of metric learning. For the negative pairs with different labels, the negative contrastive loss $\mathcal{L}_{local}^{neg}$  is denoted by 
\begin{footnotesize}
    \begin{eqnarray}
    \frac{1}{2} \max [0, \xi - \|\frac{1}{T} \sum_{i=1}^{T} \mathbb{E}[M_{\Theta}(\mathbf{z}_{n_{i}})])-\frac{1}{T} \sum_{j=1}^{T} \mathbb{E}[M_{\Theta}(\mathbf{z}_{q_{j}})])\|_{2}^{2}].\quad
\end{eqnarray}
\end{footnotesize}
As a result, a mean embedding-based contrastive loss with the view of local alignment can be calculated as
\begin{equation}
    \mathcal{L}_{local} = \mathcal{L}_{local}^{pos} + \mathcal{L}
_{local}^{neg}.
\end{equation}
Instead, we can observe from Figure \ref{p-mmd_visualized} that our proposed method induces a level-2 kernel-based MMD with empirical estimation for probabilistic embeddings. Specifically, our proposed scheme can preserve higher moments of a probabilistic embedding via nonlinear level-1 kernel (see the fourth component in Figure \ref{p-mmd_visualized}). Moreover, by introducing a level-2 kernel, the similarities between probabilistic embeddings also can be measured based on their own moment information (see the last component in Figure \ref{p-mmd_visualized}). Benefiting from these virtues, the proposed probabilistic MMD can accurately capture the discrepancy between mixture distributions via an extended empirical MMD fashion.

Here,  we validate the effectiveness of different schemes on the NKI task of Epithelium Stroma classification in each aligned view. The experimental settings are similar for different methods. The experimental results can be found in Figure \ref{parameter_modification}. As we can see, our proposed method  achieves consistent improvements in each alignment method with different Monte Carlo samples, which may be reasonable as the kernel mean representation can preserve
many statistical components due to the injective property. Second, when the number of MC samples is 10, we can observe an obvious margin in global alignment, which refers to the computation between mixture distributions.

\begin{figure}
    \centering
    \centerline{\includegraphics[width= 0.45\textwidth]{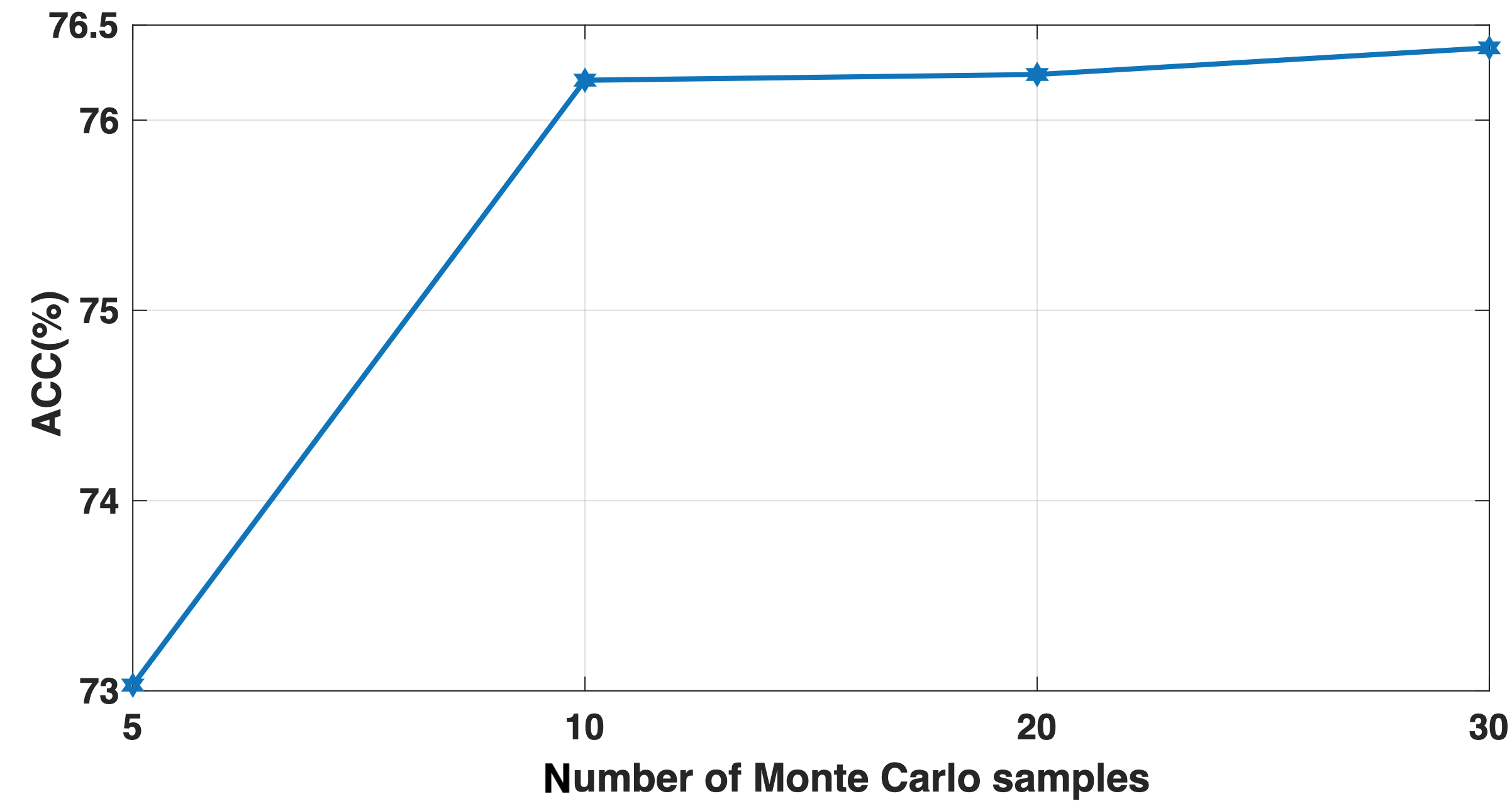}}
    \caption{The performance of our proposed model on the NKI task of Epithelium Stroma classification with different Monte Carlo samples $T$.}
    \label{mc_all}
\end{figure}

\textbf{Influence of Different Number of MC Samples\label{mceffective}.} 
It is much important to  balance the number of Monte Carlos samples and the computational efficiency. On the one hand, the property of probabilistic embeddings can be affected by the Monte Carlos sampling. On the other hand, too many Monte Carlos samples may suffer from the heavy computational cost. \citet{xiao2021bit}
suggested that the distributional property and computational cost are both acceptable for the computation of the KL divergence when the number of Monte Carlos samples is chosen appropriately, the practical performance of our proposed method needs to be explored. We conduct the experiments on the NKI task of Epithelium-Stromal classification with different Monte Carlo samples $T$. 

The results are shown in Figure \ref{mc_all}. As we can see, if the number of Monte Carlo samples is too small, it is difficult to capture the property of distribution for probabilistic embeddings. As the increase of $T$, there is an obvious improvement for our proposed method. Interestingly, the performance is gradually saturated. As a result, by balancing the number of Monte Carlos samples and the computational efficiency, the number of Monte Carlos samples $T$ in each Bayesian layer is set to 10.
\begin{table*}[!t] 
\centering
\renewcommand{\arraystretch}{1.1}
\caption{Ablation study on Epithelium Stroma Classification in
Histopathological Images. PE denotes the introduction of probabilistic embeddings.} 
\begin{adjustbox}{width=0.7\textwidth}
\begin{tabular}{cccccc}
\hline
\multicolumn{2}{c}{Method}   & IHC & NKI & \multicolumn{1}{c}{VGH}  & Average (\%)  \\ 
\hline
\multicolumn{2}{c}{Baseline}  & 79.74 $\pm$ 0.15 & 74.84$\pm$ 0.13 & 84.29$\pm$ 0.12 & 79.62 \\
 \multicolumn{2}{c}{Baseline + PE} & \textbf{81.99} $\pm$ 0.12 & \textbf{74.90}$\pm$ 0.14 & \textbf{85.01}$\pm$ 0.18 & \textbf{80.63} \\ \hline
\end{tabular}
\end{adjustbox}
\label{breast_exp_ab}
\end{table*}
\subsection{Scalability to Benchmark Datasets}

Here, we introduce three DG benchmarks, namely PACS (Art:
2048 images,
Cartoon: 2344 images,
Photo: 1670 images,
Sketch: 3929 images), OfficeHome (has 15588 samples with 65 classes from four domains) and VLCS (has 10729 samples with 5 classes from four domains), for comparison. 
Compared with some large-scale benchmarks (\textit{e.g.}, DomainNet and Wilds), these three datasets are more appropriate to explore the effectiveness of different DG models under the scenario of insufficient samples. performance. 
We adopt pretrained ResNet50 as the backbone for all benchmarks.
The structure of the overall framework is similar with the model mentioned in lesion skin classification. Our proposed method as well as baseline methods are all based on  DomainBed, where the holdout fraction (the proportion of validation set) rate for DomainBed is set to 0.2 for all methods. A domain is the target domain and the remaining 
domains are the source domain for training. The testing is on the overall data of a target domain.   

Here, our proposed method is optimized by Adam optimizer with learning rate as $5\times10^{-5}$. The batch size for each source domain is 32. The training steps are set to 20000 for PACS and OfficeHome, and 2000 for VLCS. By following the SWAD framework,  the training process will be stopped for our proposed method when the validation loss increases significantly. The hyperparameters are selected in a wide range on the validation set. For the $\mathcal{L}_{local}$ and $\mathcal{L}_{global}$, the $\beta_{1}$ and the $\beta_{2}$ are set to 0.1 and 1 for all benchmark datasets.
Other
hyperparameters such as kernel function, kernel bandwidth and distance margin are similar to the
settings mentioned before.

The experimental results on PACS, OfficeHome and VLCS can be shown in Table \ref{pacs}, \ref{officehome}, and \ref{VLCS}. As we can see, 
compared with domain-invariant-based approaches (\textit{e.g.}, DANN), our proposed method has a significant improvement due to the introduction of a probabilistic framework. Our proposed method also outperforms the data augmentation-based approach (\textit{e.g.}, Mixstyle, Manifold Mixup, and CutMix).  Although data generation methods (e.g., MixStyle) can effectively tackle the insufficient sample problem via additional generative samples, the lack of effective domain-invariant learning may hamper the improvement of the performance.  
Our proposed method achieves better performances compared with feature disentanglement-based (e.g., pAdaI). Last but not least, compared with the state-of-the-art domain generalization method (e.g., MIRO),  our proposed method achieves comparable performance on small-scale benchmark datasets, which demonstrates the scalability of our proposed method in a small-data regime.
\section{Discussion}
\textbf{Limitation.} 1) \textit{MC sampling.} Our probabilistic embeddings derive from the distribution over the model weights. Similar to widely-used BNN-based models \citep{blundell2015weight,mallick2021deep,xiao2021bit}, the predictive distribution of probabilistic embeddings needs to be approximated by MC sampling. This may result in more computational consumption compared with the original deterministic model-based scheme. Although heavy computational consumption can be alleviated in our settings (i.e., DG problem in the context of small data), the computational cost needs to be reduced more on very large-scale datasets in the future. 2) \textit{Approximate posterior.} 
Although the mean-field variational inference (MFVI) (we used) is effective in rendering an approximate posterior of BNN. The potential amortization gap due to the fully factorized Gaussian assumption of MFVI would limit further performance improvement on very challenging datasets \citep{cremer2018inference}. This limitation will be explored by replacing it with more expressive approximate posteriors (such as normalizing flows) in the future.

\textbf{Tradeoff.} In this paper, we carefully make a tradeoff between \textit{effectiveness and computational complexity}.
Specifically,
due to the introduction of probabilistic embeddings for the DG problem in the context of insufficient data, the probabilistic MMD is proposed based on the level-2 kernel for more high-level statistics, which may be more complex than other baseline methods with ``vanilla” MMD.  However, it shows improved effectiveness in addressing the problem of global semantic alignment between domains, consisting of a series of probabilistic embeddings. 
Other simpler methods may not achieve the same level of performance using just the first moment (\textit{i.e.,} the mean embedding). Moreover, we adopt some strategies, such as the unbiased estimate of MMD with linear complexity and the number selection of appropriate MC sampling, to offset the extra computational complexity compared with other baseline methods.

\textbf{Usage of our proposed method.} Compared with a deterministic framework, where a source domain as a distribution consists of a set of point embeddings, a source domain generated by our probabilistic framework includes a set of distributions, $\textit{i.e.,}$ the so-called distribution over distributions or the probability of probability. The vanilla MMD may not directly cope with this situation. Instead, our proposed P-MMD leverages the level-2 kernel to measure the discrepancy between mixture distributions based on the empirical MMD framework and preserve most information
about high-level statistics. 

In this paper, we focus on the DG problem in the context of insufficient data.  Especially, for the medical imaging data, insufficient sample scenarios either exist in all source domains or in some source domains. If the number of training samples from all source domains is sufficient, it is reasonable to choose another DG method. For instance, the Domain-Net datasets for natural images have six source domains, whereas the source domain (“clipart”) with the smallest number of training samples still has 50,000 training samples. 
Moreover, our proposed method shows good scalability on benchmark datasets with a larger number of samples, although it is designed based on the DG scenario of insufficient data.

\textbf{Gains of Probabilistic Embeddings.} To further show the gain of introducing probabilistic embeddings for insufficient data, we conducted an ablation study on the Epithelium Stroma Classification and Skin Lesion Classification. We only add a Bayesian layer (for probabilistic embeddings) with ReLU layer on the bottom of a deterministic baseline model (pre-trained ResNet-18). 

The results can be found in Tables \ref{breast_exp_ab}. As we can see, consistent improvements can be observed in these small-data tasks due to the introduction of probabilistic embeddings:

\section{Conclusion} 
In this work, we address the DG problem in the context of insufficient data, which can occur in \textit{all  or some source domains}. To this end, we introduce a probabilistic framework into the DG problem to derive probabilistic embeddings (which can be adaptive to insufficient samples better compared with deterministic models) for domain-invariant learning.  Under this probabilistic framework, an extension of MMD called P-MMD is proposed for measuring the \textit{distribution over distributions}. Moreover, a probabilistic CSA loss is proposed for local alignment. Extensive experiments on insufficient cross-domain medical imaging data show the effectiveness of this method.
\backmatter

\bmhead{Acknowledgments}
This work is supported in part by Hong Kong Innovation and Technology Commission
(InnoHK Project CIMDA), the Research Grant Council (RGC) of Hong Kong through Early Career Scheme (ECS) under the Grant 21200522, CityU Applied Research Grant (ARG) 9667244, and Sichuan Science and Technology Program 2022NSFSC0551.
\section*{Data availability Statement}
For this paper only publicly available datasets were
used. The links of Epithelium Stroma Classification, Skin Lesion Classification, and Spinal Cord Gray Matter
Segmentation can be found in http://fimm.webmicroscope.net/supplements/epist\\roma, https://challenge.isic-archive.com/landing/2018/47/, http://niftyweb.cs.ucl.ac.uk/challenge/index.php.

\section*{Declarations}

\textbf{Conflict of interest.} The authors have no conflict of interest to declare
that are relevant to the content of this article.

\begin{appendices}

\section{Details of Bayesian Neural Network} \label{bnn}
For our proposed method, the Bayesian layer refers to the probabilistic extractor $Q_{\phi}$ and the probabilistic classifier $C_{\omega}$. Here,  a simple and convenient PyTorch library, namely BayesianTorch \citep{krishnan2022bayesiantorch}, is utilized to construct the Bayesian neural network. The log evidence lower bound (ELBO) cost function, i.e.,
\begin{equation}
    \mathcal{L}:= \int q_{\theta}log(y|x,w)dw - \operatorname{KL}[q_{\theta}(w)|p(w)],
\end{equation}
can be calculated automatically. By using BayesianTorch, arbitrary deterministic models can be converted into the Bayesian layers easily.  In this paper, mean-field variational inference (MFVI) \citep{graves2011practical} is adopted, where the parameters of the model are characterized by fully factorized Gaussian distribution endowed by
variational parameters $\mu$ and $\sigma$, i.e.,
\begin{equation}
    q_{\theta}(w) := \mathcal{N}(w|\mu,\sigma).
\end{equation}
By using stochastic gradient descent method with ELBO cost,
the variational distribution $q_{\theta}(w)$ as the approximation of the posterior distribution, and corresponding parameters ($\mu$ and $\sigma$) and can be learned conveniently.

For the settings of Bayesian layer, we follow the model priors with empirical Bayes using DNN (MOPED) method for the parameter settings of weights prior, each
weight is sampled from the Gaussian distribution independently \citep{krishnan2020specifying},
\begin{equation}
    w \sim \mathcal{N}(w_{\operatorname{DNN}},\delta |w_{\operatorname{DNN}}|),
\end{equation}
where $w_{\operatorname{DNN}}$ denotes the mean of prior distribution from the maximum likelihood estimates of weights from deterministic deep neural network. $\delta$, a hyperparameter, is set to the initial perturbation factor for the percentage of the pretrained deterministic weight values.
The variational layer is modeled using reparameterization trick. The MOPED can realize better training convergence for complex models \citep{krishnan2020specifying}, which is beneficial to our proposed method. In this paper,  we follow the setting in \citep{krishnan2020specifying} to set the initial perturbation factor $\delta$ for the weight to  $0.1$. 

\section{Implementation Details of Experiments} \label{details}
\subsection{Epithelium Stroma Classification}
\textbf{Implement Details.} There are two types of basic tissues, \textit{i.e.}, the epithelium and the stroma.  Due to the differences of the scanner, the staining type, and the population, the color of the background and the morphological structure among different histopathological image datasets are diverse. The extract epithelial or stromal patches are resized into $224 \times 224$. The classification objective is the Cross-entropy loss with Softmax function.  All baseline methods are trained with the same training scheme. We tune their hyperparameters in a wide range on the validation set. The testing results are reported using the best model on validation set.

\subsection{Skin Lesion Classification}
\textbf{Implement Details.} There are seven classes of skin lesions, including melanoma (\textit{mel}), melanocytic nevus (\textit{nv}), dermato broma (\textit{df}), basal cell carcinoma (\textit{bcc}) benign keratosis (\textit{bkl}), vascular
lesion (\textit{vasc}), and actinic keratosis (\textit{akiec}).
For inputs, all images are resized into $224 \times 224$ for all methods. Due to the class imbalance problem, the focal loss \citep{lin2017focal} as the classification objective is introduced for all methods.

\subsection{Spinal Cord Gray Matter Segmentation}
\textbf{Implementation Details.} 
 By following \citep{li2018domain}, the 3D MRI data are split into 2D slices in axial view. Then, these obtained 2D slices are centered cropped to $160 \times 160$ and randomly cropped to $144 \times 144$ for training. 
The 2D-Unet \citep{ronneberger2015u} is leveraged as the backbone for all methods. For our proposed method, probabilistic extractor $Q_{\phi}$ is constructed by two Bayesian-based $1\times1$ convolutional layers. The input and output channels in the first convolutional layer are both 64. After a ReLU layer, the input and output channels in the second convolutional laye are 64 and 1, respectively. The BayesianTorch can enable to convert ordinary convolutional layer
into Bayesian convolutional neural network easily. The Bayesian neural network adopts MFVI to approximate the posterior distribution of weights. The parameters of Bayesian layer are the same as aforementioned settings. The structure of the Bayesian layer in the probabilistic classifier $C_{\omega}$ is a Bayesian-based $1\times1$ convolutional layers. The input and output channels  are 64 and 1, respectively.
The construction of $C_{\omega}$ is the same as that of $Q_{\phi}$. Here, all methods adopt a two-stage scheme for coarse-to-fine segmentation, as used in \citep{li2020domain}. Specifically, we first conduct preliminary segmentation  to obtain the spinal cord area from the original 2D slice. Then, we perform elaborative segmentation on obtained spinal cord results to derive gray matter results.

The Adam optimizer is utilized with learning rate as $1\times10^{-4}$, weight decay as $1\times10^{-8}$. The batch size is 8 for each source domain. The total epochs are 200, where the learning rate will be decreased every 80 epochs with a factor of 10. Other hyperparameters such as kernel function, kernel bandwidth and distance margin are similar to the settings in skin lesion classification and epithelium-stroma classification.
The segmentation can be regarded as the pixel-level classification. For the $\mathcal{L}_{local}$ and $\mathcal{L}_{global}$,  we follow \citep{motiian2017unified} to randomly sample some positive and negative pairs from two domains such that the computational efficiency can be improved significantly. Here, we randomly sample 400 positive and 
negative pixel pairs from two domains in a mini-batch for the computation of $\mathcal{L}_{local}$, respective. By leveraging selected pixels of a domain in $\mathcal{L}_{local}$, we further utilize these pixels to calculate the $\mathcal{L}_{global}$, which may induce a more accurate measurement owing to the balanced class distribution, as well as reducing the computational cost. 

\end{appendices}

\balance



\begin{thebibliography}{78}
\ifx \bisbn   \undefined \def \bisbn  #1{ISBN #1}\fi
\ifx \binits  \undefined \def \binits#1{#1}\fi
\ifx \bauthor  \undefined \def \bauthor#1{#1}\fi
\ifx \batitle  \undefined \def \batitle#1{#1}\fi
\ifx \bjtitle  \undefined \def \bjtitle#1{#1}\fi
\ifx \bvolume  \undefined \def \bvolume#1{\textbf{#1}}\fi
\ifx \byear  \undefined \def \byear#1{#1}\fi
\ifx \bissue  \undefined \def \bissue#1{#1}\fi
\ifx \bfpage  \undefined \def \bfpage#1{#1}\fi
\ifx \blpage  \undefined \def \blpage #1{#1}\fi
\ifx \burl  \undefined \def \burl#1{\textsf{#1}}\fi
\ifx \doiurl  \undefined \def \doiurl#1{\url{https://doi.org/#1}}\fi
\ifx \betal  \undefined \def \betal{\textit{et al.}}\fi
\ifx \binstitute  \undefined \def \binstitute#1{#1}\fi
\ifx \binstitutionaled  \undefined \def \binstitutionaled#1{#1}\fi
\ifx \bctitle  \undefined \def \bctitle#1{#1}\fi
\ifx \beditor  \undefined \def \beditor#1{#1}\fi
\ifx \bpublisher  \undefined \def \bpublisher#1{#1}\fi
\ifx \bbtitle  \undefined \def \bbtitle#1{#1}\fi
\ifx \bedition  \undefined \def \bedition#1{#1}\fi
\ifx \bseriesno  \undefined \def \bseriesno#1{#1}\fi
\ifx \blocation  \undefined \def \blocation#1{#1}\fi
\ifx \bsertitle  \undefined \def \bsertitle#1{#1}\fi
\ifx \bsnm \undefined \def \bsnm#1{#1}\fi
\ifx \bsuffix \undefined \def \bsuffix#1{#1}\fi
\ifx \bparticle \undefined \def \bparticle#1{#1}\fi
\ifx \barticle \undefined \def \barticle#1{#1}\fi
\bibcommenthead
\ifx \bconfdate \undefined \def \bconfdate #1{#1}\fi
\ifx \botherref \undefined \def \botherref #1{#1}\fi
\ifx \url \undefined \def \url#1{\textsf{#1}}\fi
\ifx \bchapter \undefined \def \bchapter#1{#1}\fi
\ifx \bbook \undefined \def \bbook#1{#1}\fi
\ifx \bcomment \undefined \def \bcomment#1{#1}\fi
\ifx \oauthor \undefined \def \oauthor#1{#1}\fi
\ifx \citeauthoryear \undefined \def \citeauthoryear#1{#1}\fi
\ifx \endbibitem  \undefined \def \endbibitem {}\fi
\ifx \bconflocation  \undefined \def \bconflocation#1{#1}\fi
\ifx \arxivurl  \undefined \def \arxivurl#1{\textsf{#1}}\fi
\csname PreBibitemsHook\endcsname

\bibitem[\protect\citeauthoryear{Li et~al.}{2022}]{li2022comprehensive}
\begin{barticle}
\bauthor{\bsnm{Li}, \binits{M.}},
\bauthor{\bsnm{Huang}, \binits{B.}},
\bauthor{\bsnm{Tian}, \binits{G.}}:
\batitle{A comprehensive survey on 3d face recognition methods}.
\bjtitle{Engineering Applications of Artificial Intelligence}
\bvolume{110},
\bfpage{104669}
(\byear{2022})
\end{barticle}
\endbibitem

\bibitem[\protect\citeauthoryear{Zaidi et~al.}{2022}]{zaidi2022survey}
\begin{botherref}
\oauthor{\bsnm{Zaidi}, \binits{S.S.A.}},
\oauthor{\bsnm{Ansari}, \binits{M.S.}},
\oauthor{\bsnm{Aslam}, \binits{A.}},
\oauthor{\bsnm{Kanwal}, \binits{N.}},
\oauthor{\bsnm{Asghar}, \binits{M.}},
\oauthor{\bsnm{Lee}, \binits{B.}}:
A survey of modern deep learning based object detection models.
Digital Signal Processing,
103514
(2022)
\end{botherref}
\endbibitem

\bibitem[\protect\citeauthoryear{Mridha et~al.}{2022}]{mridha2022study}
\begin{barticle}
\bauthor{\bsnm{Mridha}, \binits{M.F.}},
\bauthor{\bsnm{Ohi}, \binits{A.Q.}},
\bauthor{\bsnm{Hamid}, \binits{M.A.}},
\bauthor{\bsnm{Monowar}, \binits{M.M.}}:
\batitle{A study on the challenges and opportunities of speech recognition for
  bengali language}.
\bjtitle{Artificial Intelligence Review}
\bvolume{55}(\bissue{4}),
\bfpage{3431}--\blpage{3455}
(\byear{2022})
\end{barticle}
\endbibitem

\bibitem[\protect\citeauthoryear{Zhou et~al.}{2022}]{zhou2022domain}
\begin{botherref}
\oauthor{\bsnm{Zhou}, \binits{K.}},
\oauthor{\bsnm{Liu}, \binits{Z.}},
\oauthor{\bsnm{Qiao}, \binits{Y.}},
\oauthor{\bsnm{Xiang}, \binits{T.}},
\oauthor{\bsnm{Loy}, \binits{C.C.}}:
Domain generalization: A survey.
IEEE Transactions on Pattern Analysis and Machine Intelligence
(2022)
\end{botherref}
\endbibitem

\bibitem[\protect\citeauthoryear{Liu et~al.}{2022}]{liu2022deep}
\begin{botherref}
\oauthor{\bsnm{Liu}, \binits{X.}},
\oauthor{\bsnm{Yoo}, \binits{C.}},
\oauthor{\bsnm{Xing}, \binits{F.}},
\oauthor{\bsnm{Oh}, \binits{H.}},
\oauthor{\bsnm{El~Fakhri}, \binits{G.}},
\oauthor{\bsnm{Kang}, \binits{J.-W.}},
\oauthor{\bsnm{Woo}, \binits{J.}}, et al.:
Deep unsupervised domain adaptation: A review of recent advances and
  perspectives.
APSIPA Transactions on Signal and Information Processing
\textbf{11}(1)
(2022)
\end{botherref}
\endbibitem

\bibitem[\protect\citeauthoryear{Qi et~al.}{2020}]{qi2020curriculum}
\begin{barticle}
\bauthor{\bsnm{Qi}, \binits{Q.}},
\bauthor{\bsnm{Lin}, \binits{X.}},
\bauthor{\bsnm{Chen}, \binits{C.}},
\bauthor{\bsnm{Xie}, \binits{W.}},
\bauthor{\bsnm{Huang}, \binits{Y.}},
\bauthor{\bsnm{Ding}, \binits{X.}},
\bauthor{\bsnm{Liu}, \binits{X.}},
\bauthor{\bsnm{Yu}, \binits{Y.}}:
\batitle{Curriculum feature alignment domain adaptation for epithelium-stroma
  classification in histopathological images}.
\bjtitle{IEEE Journal of Biomedical and Health Informatics}
\bvolume{25}(\bissue{4}),
\bfpage{1163}--\blpage{1172}
(\byear{2020})
\end{barticle}
\endbibitem

\bibitem[\protect\citeauthoryear{Yue et~al.}{2019}]{yue2019domain}
\begin{bchapter}
\bauthor{\bsnm{Yue}, \binits{X.}},
\bauthor{\bsnm{Zhang}, \binits{Y.}},
\bauthor{\bsnm{Zhao}, \binits{S.}},
\bauthor{\bsnm{Sangiovanni-Vincentelli}, \binits{A.}},
\bauthor{\bsnm{Keutzer}, \binits{K.}},
\bauthor{\bsnm{Gong}, \binits{B.}}:
\bctitle{Domain randomization and pyramid consistency: Simulation-to-real
  generalization without accessing target domain data}.
In: \bbtitle{Proceedings of the IEEE/CVF International Conference on Computer
  Vision},
pp. \bfpage{2100}--\blpage{2110}
(\byear{2019})
\end{bchapter}
\endbibitem

\bibitem[\protect\citeauthoryear{Verma et~al.}{2019}]{pmlr-v97-verma19a}
\begin{bchapter}
\bauthor{\bsnm{Verma}, \binits{V.}},
\bauthor{\bsnm{Lamb}, \binits{A.}},
\bauthor{\bsnm{Beckham}, \binits{C.}},
\bauthor{\bsnm{Najafi}, \binits{A.}},
\bauthor{\bsnm{Mitliagkas}, \binits{I.}},
\bauthor{\bsnm{Lopez-Paz}, \binits{D.}},
\bauthor{\bsnm{Bengio}, \binits{Y.}}:
\bctitle{Manifold mixup: Better representations by interpolating hidden
  states}.
In: \beditor{\bsnm{Chaudhuri}, \binits{K.}},
\beditor{\bsnm{Salakhutdinov}, \binits{R.}} (eds.)
\bbtitle{Proceedings of the 36th International Conference on Machine Learning}.
\bsertitle{Proceedings of Machine Learning Research},
vol. \bseriesno{97},
pp. \bfpage{6438}--\blpage{6447}.
\bpublisher{PMLR}, \blocation{???}
(\byear{2019}).
\burl{https://proceedings.mlr.press/v97/verma19a.html}
\end{bchapter}
\endbibitem

\bibitem[\protect\citeauthoryear{Zhou et~al.}{2021}]{zhou2021domain}
\begin{botherref}
\oauthor{\bsnm{Zhou}, \binits{K.}},
\oauthor{\bsnm{Yang}, \binits{Y.}},
\oauthor{\bsnm{Qiao}, \binits{Y.}},
\oauthor{\bsnm{Xiang}, \binits{T.}}:
Domain generalization with mixstyle.
arXiv preprint arXiv:2104.02008
(2021)
\end{botherref}
\endbibitem

\bibitem[\protect\citeauthoryear{Li et~al.}{2018}]{li2018learning}
\begin{bchapter}
\bauthor{\bsnm{Li}, \binits{D.}},
\bauthor{\bsnm{Yang}, \binits{Y.}},
\bauthor{\bsnm{Song}, \binits{Y.-Z.}},
\bauthor{\bsnm{Hospedales}, \binits{T.}}:
\bctitle{Learning to generalize: Meta-learning for domain generalization}.
In: \bbtitle{Proceedings of the AAAI Conference on Artificial Intelligence},
vol. \bseriesno{32}
(\byear{2018})
\end{bchapter}
\endbibitem

\bibitem[\protect\citeauthoryear{Kim et~al.}{2021}]{kim2021self}
\begin{bchapter}
\bauthor{\bsnm{Kim}, \binits{J.}},
\bauthor{\bsnm{Lee}, \binits{J.}},
\bauthor{\bsnm{Park}, \binits{J.}},
\bauthor{\bsnm{Min}, \binits{D.}},
\bauthor{\bsnm{Sohn}, \binits{K.}}:
\bctitle{Self-balanced learning for domain generalization}.
In: \bbtitle{2021 IEEE International Conference on Image Processing (ICIP)},
pp. \bfpage{779}--\blpage{783}
(\byear{2021}).
\bcomment{IEEE}
\end{bchapter}
\endbibitem

\bibitem[\protect\citeauthoryear{Balaji et~al.}{2019}]{balaji2019normalized}
\begin{bchapter}
\bauthor{\bsnm{Balaji}, \binits{Y.}},
\bauthor{\bsnm{Chellappa}, \binits{R.}},
\bauthor{\bsnm{Feizi}, \binits{S.}}:
\bctitle{Normalized wasserstein for mixture distributions with applications in
  adversarial learning and domain adaptation}.
In: \bbtitle{Proceedings of the IEEE/CVF International Conference on Computer
  Vision},
pp. \bfpage{6500}--\blpage{6508}
(\byear{2019})
\end{bchapter}
\endbibitem

\bibitem[\protect\citeauthoryear{Ben-David et~al.}{2006}]{ben2006analysis}
\begin{botherref}
\oauthor{\bsnm{Ben-David}, \binits{S.}},
\oauthor{\bsnm{Blitzer}, \binits{J.}},
\oauthor{\bsnm{Crammer}, \binits{K.}},
\oauthor{\bsnm{Pereira}, \binits{F.}}:
Analysis of representations for domain adaptation.
Advances in neural information processing systems
\textbf{19}
(2006)
\end{botherref}
\endbibitem

\bibitem[\protect\citeauthoryear{Motiian et~al.}{2017}]{motiian2017unified}
\begin{bchapter}
\bauthor{\bsnm{Motiian}, \binits{S.}},
\bauthor{\bsnm{Piccirilli}, \binits{M.}},
\bauthor{\bsnm{Adjeroh}, \binits{D.A.}},
\bauthor{\bsnm{Doretto}, \binits{G.}}:
\bctitle{Unified deep supervised domain adaptation and generalization}.
In: \bbtitle{Proceedings of the IEEE International Conference on Computer
  Vision},
pp. \bfpage{5715}--\blpage{5725}
(\byear{2017})
\end{bchapter}
\endbibitem

\bibitem[\protect\citeauthoryear{Dou et~al.}{2019}]{dou2019domain}
\begin{botherref}
\oauthor{\bsnm{Dou}, \binits{Q.}},
\oauthor{\bsnm{Castro}, \binits{D.}},
\oauthor{\bsnm{Kamnitsas}, \binits{K.}},
\oauthor{\bsnm{Glocker}, \binits{B.}}:
Domain generalization via model-agnostic learning of semantic features.
Advances in Neural Information Processing Systems
\textbf{32}
(2019)
\end{botherref}
\endbibitem

\bibitem[\protect\citeauthoryear{Sohn}{2016}]{sohn2016improved}
\begin{botherref}
\oauthor{\bsnm{Sohn}, \binits{K.}}:
Improved deep metric learning with multi-class n-pair loss objective.
Advances in neural information processing systems
\textbf{29}
(2016)
\end{botherref}
\endbibitem

\bibitem[\protect\citeauthoryear{Khosla et~al.}{2020}]{khosla2020supervised}
\begin{barticle}
\bauthor{\bsnm{Khosla}, \binits{P.}},
\bauthor{\bsnm{Teterwak}, \binits{P.}},
\bauthor{\bsnm{Wang}, \binits{C.}},
\bauthor{\bsnm{Sarna}, \binits{A.}},
\bauthor{\bsnm{Tian}, \binits{Y.}},
\bauthor{\bsnm{Isola}, \binits{P.}},
\bauthor{\bsnm{Maschinot}, \binits{A.}},
\bauthor{\bsnm{Liu}, \binits{C.}},
\bauthor{\bsnm{Krishnan}, \binits{D.}}:
\batitle{Supervised contrastive learning}.
\bjtitle{Advances in Neural Information Processing Systems}
\bvolume{33},
\bfpage{18661}--\blpage{18673}
(\byear{2020})
\end{barticle}
\endbibitem

\bibitem[\protect\citeauthoryear{Yao et~al.}{2022}]{yao2022pcl}
\begin{bchapter}
\bauthor{\bsnm{Yao}, \binits{X.}},
\bauthor{\bsnm{Bai}, \binits{Y.}},
\bauthor{\bsnm{Zhang}, \binits{X.}},
\bauthor{\bsnm{Zhang}, \binits{Y.}},
\bauthor{\bsnm{Sun}, \binits{Q.}},
\bauthor{\bsnm{Chen}, \binits{R.}},
\bauthor{\bsnm{Li}, \binits{R.}},
\bauthor{\bsnm{Yu}, \binits{B.}}:
\bctitle{Pcl: Proxy-based contrastive learning for domain generalization}.
In: \bbtitle{Proceedings of the IEEE/CVF Conference on Computer Vision and
  Pattern Recognition},
pp. \bfpage{7097}--\blpage{7107}
(\byear{2022})
\end{bchapter}
\endbibitem

\bibitem[\protect\citeauthoryear{Bu et~al.}{2018}]{bu2018estimation}
\begin{barticle}
\bauthor{\bsnm{Bu}, \binits{Y.}},
\bauthor{\bsnm{Zou}, \binits{S.}},
\bauthor{\bsnm{Liang}, \binits{Y.}},
\bauthor{\bsnm{Veeravalli}, \binits{V.V.}}:
\batitle{Estimation of kl divergence: Optimal minimax rate}.
\bjtitle{IEEE Transactions on Information Theory}
\bvolume{64}(\bissue{4}),
\bfpage{2648}--\blpage{2674}
(\byear{2018})
\end{barticle}
\endbibitem

\bibitem[\protect\citeauthoryear{Lee et~al.}{2022}]{lee2022deep}
\begin{botherref}
\oauthor{\bsnm{Lee}, \binits{J.}},
\oauthor{\bsnm{Liu}, \binits{C.}},
\oauthor{\bsnm{Kim}, \binits{J.}},
\oauthor{\bsnm{Chen}, \binits{Z.}},
\oauthor{\bsnm{Sun}, \binits{Y.}},
\oauthor{\bsnm{Rogers}, \binits{J.R.}},
\oauthor{\bsnm{Chung}, \binits{W.K.}},
\oauthor{\bsnm{Weng}, \binits{C.}}:
Deep learning for rare disease: A scoping review.
medRxiv
(2022)
\end{botherref}
\endbibitem

\bibitem[\protect\citeauthoryear{Johnson and Louis}{2022}]{johnson2022does}
\begin{barticle}
\bauthor{\bsnm{Johnson}, \binits{J.D.}},
\bauthor{\bsnm{Louis}, \binits{J.M.}}:
\batitle{Does race or ethnicity play a role in the origin, pathophysiology, and
  outcomes of preeclampsia? an expert review of the literature}.
\bjtitle{American journal of obstetrics and gynecology}
\bvolume{226}(\bissue{2}),
\bfpage{876}--\blpage{885}
(\byear{2022})
\end{barticle}
\endbibitem

\bibitem[\protect\citeauthoryear{Gurdasani
  et~al.}{2019}]{gurdasani2019genomics}
\begin{barticle}
\bauthor{\bsnm{Gurdasani}, \binits{D.}},
\bauthor{\bsnm{Barroso}, \binits{I.}},
\bauthor{\bsnm{Zeggini}, \binits{E.}},
\bauthor{\bsnm{Sandhu}, \binits{M.S.}}:
\batitle{Genomics of disease risk in globally diverse populations}.
\bjtitle{Nature Reviews Genetics}
\bvolume{20}(\bissue{9}),
\bfpage{520}--\blpage{535}
(\byear{2019})
\end{barticle}
\endbibitem

\bibitem[\protect\citeauthoryear{Can and Ersoy}{2021}]{can2021privacy}
\begin{barticle}
\bauthor{\bsnm{Can}, \binits{Y.S.}},
\bauthor{\bsnm{Ersoy}, \binits{C.}}:
\batitle{Privacy-preserving federated deep learning for wearable iot-based
  biomedical monitoring}.
\bjtitle{ACM Transactions on Internet Technology (TOIT)}
\bvolume{21}(\bissue{1}),
\bfpage{1}--\blpage{17}
(\byear{2021})
\end{barticle}
\endbibitem

\bibitem[\protect\citeauthoryear{Graves}{2011}]{graves2011practical}
\begin{botherref}
\oauthor{\bsnm{Graves}, \binits{A.}}:
Practical variational inference for neural networks.
Advances in neural information processing systems
\textbf{24}
(2011)
\end{botherref}
\endbibitem

\bibitem[\protect\citeauthoryear{Li et~al.}{2018}]{li2018domain}
\begin{bchapter}
\bauthor{\bsnm{Li}, \binits{H.}},
\bauthor{\bsnm{Pan}, \binits{S.J.}},
\bauthor{\bsnm{Wang}, \binits{S.}},
\bauthor{\bsnm{Kot}, \binits{A.C.}}:
\bctitle{Domain generalization with adversarial feature learning}.
In: \bbtitle{Proceedings of the IEEE Conference on Computer Vision and Pattern
  Recognition},
pp. \bfpage{5400}--\blpage{5409}
(\byear{2018})
\end{bchapter}
\endbibitem

\bibitem[\protect\citeauthoryear{Gong et~al.}{2019}]{gong2019dlow}
\begin{bchapter}
\bauthor{\bsnm{Gong}, \binits{R.}},
\bauthor{\bsnm{Li}, \binits{W.}},
\bauthor{\bsnm{Chen}, \binits{Y.}},
\bauthor{\bsnm{Gool}, \binits{L.V.}}:
\bctitle{Dlow: Domain flow for adaptation and generalization}.
In: \bbtitle{Proceedings of the IEEE/CVF Conference on Computer Vision and
  Pattern Recognition},
pp. \bfpage{2477}--\blpage{2486}
(\byear{2019})
\end{bchapter}
\endbibitem

\bibitem[\protect\citeauthoryear{Xiao et~al.}{2021}]{xiao2021bit}
\begin{bchapter}
\bauthor{\bsnm{Xiao}, \binits{Z.}},
\bauthor{\bsnm{Shen}, \binits{J.}},
\bauthor{\bsnm{Zhen}, \binits{X.}},
\bauthor{\bsnm{Shao}, \binits{L.}},
\bauthor{\bsnm{Snoek}, \binits{C.}}:
\bctitle{A bit more bayesian: Domain-invariant learning with uncertainty}.
In: \bbtitle{International Conference on Machine Learning},
pp. \bfpage{11351}--\blpage{11361}
(\byear{2021}).
\bcomment{PMLR}
\end{bchapter}
\endbibitem

\bibitem[\protect\citeauthoryear{Liu et~al.}{2021}]{liu2021feddg}
\begin{bchapter}
\bauthor{\bsnm{Liu}, \binits{Q.}},
\bauthor{\bsnm{Chen}, \binits{C.}},
\bauthor{\bsnm{Qin}, \binits{J.}},
\bauthor{\bsnm{Dou}, \binits{Q.}},
\bauthor{\bsnm{Heng}, \binits{P.-A.}}:
\bctitle{Feddg: Federated domain generalization on medical image segmentation
  via episodic learning in continuous frequency space}.
In: \bbtitle{Proceedings of the IEEE/CVF Conference on Computer Vision and
  Pattern Recognition},
pp. \bfpage{1013}--\blpage{1023}
(\byear{2021})
\end{bchapter}
\endbibitem

\bibitem[\protect\citeauthoryear{Li et~al.}{2022}]{li2022domain}
\begin{barticle}
\bauthor{\bsnm{Li}, \binits{C.}},
\bauthor{\bsnm{Lin}, \binits{X.}},
\bauthor{\bsnm{Mao}, \binits{Y.}},
\bauthor{\bsnm{Lin}, \binits{W.}},
\bauthor{\bsnm{Qi}, \binits{Q.}},
\bauthor{\bsnm{Ding}, \binits{X.}},
\bauthor{\bsnm{Huang}, \binits{Y.}},
\bauthor{\bsnm{Liang}, \binits{D.}},
\bauthor{\bsnm{Yu}, \binits{Y.}}:
\batitle{Domain generalization on medical imaging classification using episodic
  training with task augmentation}.
\bjtitle{Computers in Biology and Medicine}
\bvolume{141},
\bfpage{105144}
(\byear{2022})
\end{barticle}
\endbibitem

\bibitem[\protect\citeauthoryear{Li et~al.}{2020}]{li2020domain}
\begin{barticle}
\bauthor{\bsnm{Li}, \binits{H.}},
\bauthor{\bsnm{Wang}, \binits{Y.}},
\bauthor{\bsnm{Wan}, \binits{R.}},
\bauthor{\bsnm{Wang}, \binits{S.}},
\bauthor{\bsnm{Li}, \binits{T.-Q.}},
\bauthor{\bsnm{Kot}, \binits{A.}}:
\batitle{Domain generalization for medical imaging classification with
  linear-dependency regularization}.
\bjtitle{Advances in Neural Information Processing Systems}
\bvolume{33},
\bfpage{3118}--\blpage{3129}
(\byear{2020})
\end{barticle}
\endbibitem

\bibitem[\protect\citeauthoryear{Kingma et~al.}{2015}]{kingma2015variational}
\begin{botherref}
\oauthor{\bsnm{Kingma}, \binits{D.P.}},
\oauthor{\bsnm{Salimans}, \binits{T.}},
\oauthor{\bsnm{Welling}, \binits{M.}}:
Variational dropout and the local reparameterization trick.
Advances in neural information processing systems
\textbf{28}
(2015)
\end{botherref}
\endbibitem

\bibitem[\protect\citeauthoryear{Gal and Ghahramani}{2016}]{gal2016dropout}
\begin{bchapter}
\bauthor{\bsnm{Gal}, \binits{Y.}},
\bauthor{\bsnm{Ghahramani}, \binits{Z.}}:
\bctitle{Dropout as a bayesian approximation: Representing model uncertainty in
  deep learning}.
In: \bbtitle{International Conference on Machine Learning},
pp. \bfpage{1050}--\blpage{1059}
(\byear{2016}).
\bcomment{PMLR}
\end{bchapter}
\endbibitem

\bibitem[\protect\citeauthoryear{Blundell et~al.}{2015}]{blundell2015weight}
\begin{bchapter}
\bauthor{\bsnm{Blundell}, \binits{C.}},
\bauthor{\bsnm{Cornebise}, \binits{J.}},
\bauthor{\bsnm{Kavukcuoglu}, \binits{K.}},
\bauthor{\bsnm{Wierstra}, \binits{D.}}:
\bctitle{Weight uncertainty in neural network}.
In: \bbtitle{International Conference on Machine Learning},
pp. \bfpage{1613}--\blpage{1622}
(\byear{2015}).
\bcomment{PMLR}
\end{bchapter}
\endbibitem

\bibitem[\protect\citeauthoryear{Neal}{2012}]{neal2012bayesian}
\begin{bbook}
\bauthor{\bsnm{Neal}, \binits{R.M.}}:
\bbtitle{Bayesian Learning for Neural Networks}
vol. \bseriesno{118}.
\bpublisher{Springer}, \blocation{???}
(\byear{2012})
\end{bbook}
\endbibitem

\bibitem[\protect\citeauthoryear{Wilson and
  Izmailov}{2020}]{wilson2020bayesian}
\begin{barticle}
\bauthor{\bsnm{Wilson}, \binits{A.G.}},
\bauthor{\bsnm{Izmailov}, \binits{P.}}:
\batitle{Bayesian deep learning and a probabilistic perspective of
  generalization}.
\bjtitle{Advances in neural information processing systems}
\bvolume{33},
\bfpage{4697}--\blpage{4708}
(\byear{2020})
\end{barticle}
\endbibitem

\bibitem[\protect\citeauthoryear{Mallick et~al.}{2021}]{mallick2021deep}
\begin{bchapter}
\bauthor{\bsnm{Mallick}, \binits{A.}},
\bauthor{\bsnm{Dwivedi}, \binits{C.}},
\bauthor{\bsnm{Kailkhura}, \binits{B.}},
\bauthor{\bsnm{Joshi}, \binits{G.}},
\bauthor{\bsnm{Han}, \binits{T.Y.-J.}}:
\bctitle{Deep kernels with probabilistic embeddings for small-data learning}.
In: \bbtitle{Uncertainty in Artificial Intelligence},
pp. \bfpage{918}--\blpage{928}
(\byear{2021}).
\bcomment{PMLR}
\end{bchapter}
\endbibitem

\bibitem[\protect\citeauthoryear{Blei et~al.}{2017}]{blei2017variational}
\begin{barticle}
\bauthor{\bsnm{Blei}, \binits{D.M.}},
\bauthor{\bsnm{Kucukelbir}, \binits{A.}},
\bauthor{\bsnm{McAuliffe}, \binits{J.D.}}:
\batitle{Variational inference: A review for statisticians}.
\bjtitle{Journal of the American statistical Association}
\bvolume{112}(\bissue{518}),
\bfpage{859}--\blpage{877}
(\byear{2017})
\end{barticle}
\endbibitem

\bibitem[\protect\citeauthoryear{Krishnan
  et~al.}{2020}]{krishnan2020specifying}
\begin{bchapter}
\bauthor{\bsnm{Krishnan}, \binits{R.}},
\bauthor{\bsnm{Subedar}, \binits{M.}},
\bauthor{\bsnm{Tickoo}, \binits{O.}}:
\bctitle{Specifying weight priors in bayesian deep neural networks with
  empirical bayes}.
In: \bbtitle{Proceedings of the AAAI Conference on Artificial Intelligence},
vol. \bseriesno{34},
pp. \bfpage{4477}--\blpage{4484}
(\byear{2020})
\end{bchapter}
\endbibitem

\bibitem[\protect\citeauthoryear{Nguyen et~al.}{2017}]{nguyen2017mixture}
\begin{botherref}
\oauthor{\bsnm{Nguyen}, \binits{D.Q.}},
\oauthor{\bsnm{Nguyen}, \binits{D.Q.}},
\oauthor{\bsnm{Modi}, \binits{A.}},
\oauthor{\bsnm{Thater}, \binits{S.}},
\oauthor{\bsnm{Pinkal}, \binits{M.}}:
A mixture model for learning multi-sense word embeddings.
arXiv preprint arXiv:1706.05111
(2017)
\end{botherref}
\endbibitem

\bibitem[\protect\citeauthoryear{Park et~al.}{2022}]{park2022probabilistic}
\begin{bchapter}
\bauthor{\bsnm{Park}, \binits{J.}},
\bauthor{\bsnm{Lee}, \binits{J.}},
\bauthor{\bsnm{Kim}, \binits{I.-J.}},
\bauthor{\bsnm{Sohn}, \binits{K.}}:
\bctitle{Probabilistic representations for video contrastive learning}.
In: \bbtitle{Proceedings of the IEEE/CVF Conference on Computer Vision and
  Pattern Recognition},
pp. \bfpage{14711}--\blpage{14721}
(\byear{2022})
\end{bchapter}
\endbibitem

\bibitem[\protect\citeauthoryear{Oh et~al.}{2018}]{oh2018modeling}
\begin{botherref}
\oauthor{\bsnm{Oh}, \binits{S.J.}},
\oauthor{\bsnm{Murphy}, \binits{K.}},
\oauthor{\bsnm{Pan}, \binits{J.}},
\oauthor{\bsnm{Roth}, \binits{J.}},
\oauthor{\bsnm{Schroff}, \binits{F.}},
\oauthor{\bsnm{Gallagher}, \binits{A.}}:
Modeling uncertainty with hedged instance embedding.
arXiv preprint arXiv:1810.00319
(2018)
\end{botherref}
\endbibitem

\bibitem[\protect\citeauthoryear{Shi and Jain}{2019}]{shi2019probabilistic}
\begin{bchapter}
\bauthor{\bsnm{Shi}, \binits{Y.}},
\bauthor{\bsnm{Jain}, \binits{A.K.}}:
\bctitle{Probabilistic face embeddings}.
In: \bbtitle{Proceedings of the IEEE/CVF International Conference on Computer
  Vision},
pp. \bfpage{6902}--\blpage{6911}
(\byear{2019})
\end{bchapter}
\endbibitem

\bibitem[\protect\citeauthoryear{Chang et~al.}{2020}]{chang2020data}
\begin{bchapter}
\bauthor{\bsnm{Chang}, \binits{J.}},
\bauthor{\bsnm{Lan}, \binits{Z.}},
\bauthor{\bsnm{Cheng}, \binits{C.}},
\bauthor{\bsnm{Wei}, \binits{Y.}}:
\bctitle{Data uncertainty learning in face recognition}.
In: \bbtitle{Proceedings of the IEEE/CVF Conference on Computer Vision and
  Pattern Recognition},
pp. \bfpage{5710}--\blpage{5719}
(\byear{2020})
\end{bchapter}
\endbibitem

\bibitem[\protect\citeauthoryear{Silnova
  et~al.}{2020}]{silnova2020probabilistic}
\begin{botherref}
\oauthor{\bsnm{Silnova}, \binits{A.}},
\oauthor{\bsnm{Br{\"u}mmer}, \binits{N.}},
\oauthor{\bsnm{Rohdin}, \binits{J.}},
\oauthor{\bsnm{Stafylakis}, \binits{T.}},
\oauthor{\bsnm{Burget}, \binits{L.}}:
Probabilistic embeddings for speaker diarization.
arXiv preprint arXiv:2004.04096
(2020)
\end{botherref}
\endbibitem

\bibitem[\protect\citeauthoryear{Sun et~al.}{2020}]{sun2020view}
\begin{bchapter}
\bauthor{\bsnm{Sun}, \binits{J.J.}},
\bauthor{\bsnm{Zhao}, \binits{J.}},
\bauthor{\bsnm{Chen}, \binits{L.-C.}},
\bauthor{\bsnm{Schroff}, \binits{F.}},
\bauthor{\bsnm{Adam}, \binits{H.}},
\bauthor{\bsnm{Liu}, \binits{T.}}:
\bctitle{View-invariant probabilistic embedding for human pose}.
In: \bbtitle{Computer Vision--ECCV 2020: 16th European Conference, Glasgow, UK,
  August 23--28, 2020, Proceedings, Part V 16},
pp. \bfpage{53}--\blpage{70}
(\byear{2020}).
\bcomment{Springer}
\end{bchapter}
\endbibitem

\bibitem[\protect\citeauthoryear{Chun et~al.}{2021}]{chun2021probabilistic}
\begin{bchapter}
\bauthor{\bsnm{Chun}, \binits{S.}},
\bauthor{\bsnm{Oh}, \binits{S.J.}},
\bauthor{\bsnm{De~Rezende}, \binits{R.S.}},
\bauthor{\bsnm{Kalantidis}, \binits{Y.}},
\bauthor{\bsnm{Larlus}, \binits{D.}}:
\bctitle{Probabilistic embeddings for cross-modal retrieval}.
In: \bbtitle{Proceedings of the IEEE/CVF Conference on Computer Vision and
  Pattern Recognition},
pp. \bfpage{8415}--\blpage{8424}
(\byear{2021})
\end{bchapter}
\endbibitem

\bibitem[\protect\citeauthoryear{Neculai
  et~al.}{2022}]{neculai2022probabilistic}
\begin{bchapter}
\bauthor{\bsnm{Neculai}, \binits{A.}},
\bauthor{\bsnm{Chen}, \binits{Y.}},
\bauthor{\bsnm{Akata}, \binits{Z.}}:
\bctitle{Probabilistic compositional embeddings for multimodal image
  retrieval}.
In: \bbtitle{Proceedings of the IEEE/CVF Conference on Computer Vision and
  Pattern Recognition},
pp. \bfpage{4547}--\blpage{4557}
(\byear{2022})
\end{bchapter}
\endbibitem

\bibitem[\protect\citeauthoryear{Chun}{2023}]{chun2023improved}
\begin{botherref}
\oauthor{\bsnm{Chun}, \binits{S.}}:
Improved probabilistic image-text representations.
arXiv preprint arXiv:2305.18171
(2023)
\end{botherref}
\endbibitem

\bibitem[\protect\citeauthoryear{Long et~al.}{2017}]{long2017deep}
\begin{bchapter}
\bauthor{\bsnm{Long}, \binits{M.}},
\bauthor{\bsnm{Zhu}, \binits{H.}},
\bauthor{\bsnm{Wang}, \binits{J.}},
\bauthor{\bsnm{Jordan}, \binits{M.I.}}:
\bctitle{Deep transfer learning with joint adaptation networks}.
In: \bbtitle{International Conference on Machine Learning},
pp. \bfpage{2208}--\blpage{2217}
(\byear{2017}).
\bcomment{PMLR}
\end{bchapter}
\endbibitem

\bibitem[\protect\citeauthoryear{Borgwardt
  et~al.}{2006}]{borgwardt2006integrating}
\begin{barticle}
\bauthor{\bsnm{Borgwardt}, \binits{K.M.}},
\bauthor{\bsnm{Gretton}, \binits{A.}},
\bauthor{\bsnm{Rasch}, \binits{M.J.}},
\bauthor{\bsnm{Kriegel}, \binits{H.-P.}},
\bauthor{\bsnm{Sch{\"o}lkopf}, \binits{B.}},
\bauthor{\bsnm{Smola}, \binits{A.J.}}:
\batitle{Integrating structured biological data by kernel maximum mean
  discrepancy}.
\bjtitle{Bioinformatics}
\bvolume{22}(\bissue{14}),
\bfpage{49}--\blpage{57}
(\byear{2006})
\end{barticle}
\endbibitem

\bibitem[\protect\citeauthoryear{Gretton et~al.}{2012}]{gretton2012kernel}
\begin{barticle}
\bauthor{\bsnm{Gretton}, \binits{A.}},
\bauthor{\bsnm{Borgwardt}, \binits{K.M.}},
\bauthor{\bsnm{Rasch}, \binits{M.J.}},
\bauthor{\bsnm{Sch{\"o}lkopf}, \binits{B.}},
\bauthor{\bsnm{Smola}, \binits{A.}}:
\batitle{A kernel two-sample test}.
\bjtitle{The Journal of Machine Learning Research}
\bvolume{13}(\bissue{1}),
\bfpage{723}--\blpage{773}
(\byear{2012})
\end{barticle}
\endbibitem

\bibitem[\protect\citeauthoryear{Hu et~al.}{2020}]{hu2020domain}
\begin{bchapter}
\bauthor{\bsnm{Hu}, \binits{S.}},
\bauthor{\bsnm{Zhang}, \binits{K.}},
\bauthor{\bsnm{Chen}, \binits{Z.}},
\bauthor{\bsnm{Chan}, \binits{L.}}:
\bctitle{Domain generalization via multidomain discriminant analysis}.
In: \bbtitle{Uncertainty in Artificial Intelligence},
pp. \bfpage{292}--\blpage{302}
(\byear{2020}).
\bcomment{PMLR}
\end{bchapter}
\endbibitem

\bibitem[\protect\citeauthoryear{Yoshikawa et~al.}{2014}]{yoshikawa2014latent}
\begin{botherref}
\oauthor{\bsnm{Yoshikawa}, \binits{Y.}},
\oauthor{\bsnm{Iwata}, \binits{T.}},
\oauthor{\bsnm{Sawada}, \binits{H.}}:
Latent support measure machines for bag-of-words data classification.
Advances in neural information processing systems
\textbf{27}
(2014)
\end{botherref}
\endbibitem

\bibitem[\protect\citeauthoryear{Muandet et~al.}{2012}]{muandet2012learning}
\begin{botherref}
\oauthor{\bsnm{Muandet}, \binits{K.}},
\oauthor{\bsnm{Fukumizu}, \binits{K.}},
\oauthor{\bsnm{Dinuzzo}, \binits{F.}},
\oauthor{\bsnm{Sch{\"o}lkopf}, \binits{B.}}:
Learning from distributions via support measure machines.
Advances in neural information processing systems
\textbf{25}
(2012)
\end{botherref}
\endbibitem

\bibitem[\protect\citeauthoryear{Berlinet and
  Thomas-Agnan}{2011}]{berlinet2011reproducing}
\begin{bbook}
\bauthor{\bsnm{Berlinet}, \binits{A.}},
\bauthor{\bsnm{Thomas-Agnan}, \binits{C.}}:
\bbtitle{Reproducing Kernel Hilbert Spaces in Probability and Statistics}.
\bpublisher{Springer}, \blocation{???}
(\byear{2011})
\end{bbook}
\endbibitem

\bibitem[\protect\citeauthoryear{Muandet et~al.}{2017}]{muandet2017kernel}
\begin{barticle}
\bauthor{\bsnm{Muandet}, \binits{K.}},
\bauthor{\bsnm{Fukumizu}, \binits{K.}},
\bauthor{\bsnm{Sriperumbudur}, \binits{B.}},
\bauthor{\bsnm{Sch{\"o}lkopf}, \binits{B.}}, \betal:
\batitle{Kernel mean embedding of distributions: A review and beyond}.
\bjtitle{Foundations and Trends{\textregistered} in Machine Learning}
\bvolume{10}(\bissue{1-2}),
\bfpage{1}--\blpage{141}
(\byear{2017})
\end{barticle}
\endbibitem

\bibitem[\protect\citeauthoryear{Cha et~al.}{2021}]{cha2021swad}
\begin{barticle}
\bauthor{\bsnm{Cha}, \binits{J.}},
\bauthor{\bsnm{Chun}, \binits{S.}},
\bauthor{\bsnm{Lee}, \binits{K.}},
\bauthor{\bsnm{Cho}, \binits{H.-C.}},
\bauthor{\bsnm{Park}, \binits{S.}},
\bauthor{\bsnm{Lee}, \binits{Y.}},
\bauthor{\bsnm{Park}, \binits{S.}}:
\batitle{Swad: Domain generalization by seeking flat minima}.
\bjtitle{Advances in Neural Information Processing Systems}
\bvolume{34},
\bfpage{22405}--\blpage{22418}
(\byear{2021})
\end{barticle}
\endbibitem

\bibitem[\protect\citeauthoryear{Wang et~al.}{2021}]{wang2021embracing}
\begin{bchapter}
\bauthor{\bsnm{Wang}, \binits{Y.}},
\bauthor{\bsnm{Li}, \binits{H.}},
\bauthor{\bsnm{Chau}, \binits{L.-p.}},
\bauthor{\bsnm{Kot}, \binits{A.C.}}:
\bctitle{Embracing the dark knowledge: Domain generalization using regularized
  knowledge distillation}.
In: \bbtitle{Proceedings of the 29th ACM International Conference on
  Multimedia},
pp. \bfpage{2595}--\blpage{2604}
(\byear{2021})
\end{bchapter}
\endbibitem

\bibitem[\protect\citeauthoryear{Chu et~al.}{2022}]{chu2022dna}
\begin{bchapter}
\bauthor{\bsnm{Chu}, \binits{X.}},
\bauthor{\bsnm{Jin}, \binits{Y.}},
\bauthor{\bsnm{Zhu}, \binits{W.}},
\bauthor{\bsnm{Wang}, \binits{Y.}},
\bauthor{\bsnm{Wang}, \binits{X.}},
\bauthor{\bsnm{Zhang}, \binits{S.}},
\bauthor{\bsnm{Mei}, \binits{H.}}:
\bctitle{Dna: Domain generalization with diversified neural averaging}.
In: \bbtitle{International Conference on Machine Learning},
pp. \bfpage{4010}--\blpage{4034}
(\byear{2022}).
\bcomment{PMLR}
\end{bchapter}
\endbibitem

\bibitem[\protect\citeauthoryear{Li et~al.}{2022}]{li2022uncertainty}
\begin{botherref}
\oauthor{\bsnm{Li}, \binits{X.}},
\oauthor{\bsnm{Dai}, \binits{Y.}},
\oauthor{\bsnm{Ge}, \binits{Y.}},
\oauthor{\bsnm{Liu}, \binits{J.}},
\oauthor{\bsnm{Shan}, \binits{Y.}},
\oauthor{\bsnm{Duan}, \binits{L.-Y.}}:
Uncertainty modeling for out-of-distribution generalization.
arXiv preprint arXiv:2202.03958
(2022)
\end{botherref}
\endbibitem

\bibitem[\protect\citeauthoryear{Cha et~al.}{2022}]{cha2022domain}
\begin{bchapter}
\bauthor{\bsnm{Cha}, \binits{J.}},
\bauthor{\bsnm{Lee}, \binits{K.}},
\bauthor{\bsnm{Park}, \binits{S.}},
\bauthor{\bsnm{Chun}, \binits{S.}}:
\bctitle{Domain generalization by mutual-information regularization with
  pre-trained models}.
In: \bbtitle{Computer Vision--ECCV 2022: 17th European Conference, Tel Aviv,
  Israel, October 23--27, 2022, Proceedings, Part XXIII},
pp. \bfpage{440}--\blpage{457}
(\byear{2022}).
\bcomment{Springer}
\end{bchapter}
\endbibitem

\bibitem[\protect\citeauthoryear{Ronneberger et~al.}{2015}]{ronneberger2015u}
\begin{bchapter}
\bauthor{\bsnm{Ronneberger}, \binits{O.}},
\bauthor{\bsnm{Fischer}, \binits{P.}},
\bauthor{\bsnm{Brox}, \binits{T.}}:
\bctitle{U-net: Convolutional networks for biomedical image segmentation}.
In: \bbtitle{International Conference on Medical Image Computing and
  Computer-assisted Intervention},
pp. \bfpage{234}--\blpage{241}
(\byear{2015}).
\bcomment{Springer}
\end{bchapter}
\endbibitem

\bibitem[\protect\citeauthoryear{Huang et~al.}{2020}]{huang2020self}
\begin{bchapter}
\bauthor{\bsnm{Huang}, \binits{Z.}},
\bauthor{\bsnm{Wang}, \binits{H.}},
\bauthor{\bsnm{Xing}, \binits{E.P.}},
\bauthor{\bsnm{Huang}, \binits{D.}}:
\bctitle{Self-challenging improves cross-domain generalization}.
In: \bbtitle{European Conference on Computer Vision},
pp. \bfpage{124}--\blpage{140}
(\byear{2020}).
\bcomment{Springer}
\end{bchapter}
\endbibitem

\bibitem[\protect\citeauthoryear{Zhou et~al.}{2020}]{zhou2020learning}
\begin{bchapter}
\bauthor{\bsnm{Zhou}, \binits{K.}},
\bauthor{\bsnm{Yang}, \binits{Y.}},
\bauthor{\bsnm{Hospedales}, \binits{T.}},
\bauthor{\bsnm{Xiang}, \binits{T.}}:
\bctitle{Learning to generate novel domains for domain generalization}.
In: \bbtitle{European Conference on Computer Vision},
pp. \bfpage{561}--\blpage{578}
(\byear{2020}).
\bcomment{Springer}
\end{bchapter}
\endbibitem

\bibitem[\protect\citeauthoryear{Mahajan et~al.}{2021}]{mahajan2021domain}
\begin{bchapter}
\bauthor{\bsnm{Mahajan}, \binits{D.}},
\bauthor{\bsnm{Tople}, \binits{S.}},
\bauthor{\bsnm{Sharma}, \binits{A.}}:
\bctitle{Domain generalization using causal matching}.
In: \bbtitle{International Conference on Machine Learning},
pp. \bfpage{7313}--\blpage{7324}
(\byear{2021}).
\bcomment{PMLR}
\end{bchapter}
\endbibitem

\bibitem[\protect\citeauthoryear{Nuriel et~al.}{2021}]{nuriel2021permuted}
\begin{bchapter}
\bauthor{\bsnm{Nuriel}, \binits{O.}},
\bauthor{\bsnm{Benaim}, \binits{S.}},
\bauthor{\bsnm{Wolf}, \binits{L.}}:
\bctitle{Permuted adain: Reducing the bias towards global statistics in image
  classification}.
In: \bbtitle{Proceedings of the IEEE/CVF Conference on Computer Vision and
  Pattern Recognition},
pp. \bfpage{9482}--\blpage{9491}
(\byear{2021})
\end{bchapter}
\endbibitem

\bibitem[\protect\citeauthoryear{Nam et~al.}{2021}]{nam2021reducing}
\begin{bchapter}
\bauthor{\bsnm{Nam}, \binits{H.}},
\bauthor{\bsnm{Lee}, \binits{H.}},
\bauthor{\bsnm{Park}, \binits{J.}},
\bauthor{\bsnm{Yoon}, \binits{W.}},
\bauthor{\bsnm{Yoo}, \binits{D.}}:
\bctitle{Reducing domain gap by reducing style bias}.
In: \bbtitle{Proceedings of the IEEE/CVF Conference on Computer Vision and
  Pattern Recognition},
pp. \bfpage{8690}--\blpage{8699}
(\byear{2021})
\end{bchapter}
\endbibitem

\bibitem[\protect\citeauthoryear{Vapnik}{1999}]{vapnik1999overview}
\begin{barticle}
\bauthor{\bsnm{Vapnik}, \binits{V.N.}}:
\batitle{An overview of statistical learning theory}.
\bjtitle{IEEE transactions on neural networks}
\bvolume{10}(\bissue{5}),
\bfpage{988}--\blpage{999}
(\byear{1999})
\end{barticle}
\endbibitem

\bibitem[\protect\citeauthoryear{Ganin et~al.}{2016}]{ganin2016domain}
\begin{barticle}
\bauthor{\bsnm{Ganin}, \binits{Y.}},
\bauthor{\bsnm{Ustinova}, \binits{E.}},
\bauthor{\bsnm{Ajakan}, \binits{H.}},
\bauthor{\bsnm{Germain}, \binits{P.}},
\bauthor{\bsnm{Larochelle}, \binits{H.}},
\bauthor{\bsnm{Laviolette}, \binits{F.}},
\bauthor{\bsnm{Marchand}, \binits{M.}},
\bauthor{\bsnm{Lempitsky}, \binits{V.}}:
\batitle{Domain-adversarial training of neural networks}.
\bjtitle{The journal of machine learning research}
\bvolume{17}(\bissue{1}),
\bfpage{2096}--\blpage{2030}
(\byear{2016})
\end{barticle}
\endbibitem

\bibitem[\protect\citeauthoryear{Sagawa
  et~al.}{2019}]{sagawa2019distributionally}
\begin{botherref}
\oauthor{\bsnm{Sagawa}, \binits{S.}},
\oauthor{\bsnm{Koh}, \binits{P.W.}},
\oauthor{\bsnm{Hashimoto}, \binits{T.B.}},
\oauthor{\bsnm{Liang}, \binits{P.}}:
Distributionally robust neural networks for group shifts: On the importance of
  regularization for worst-case generalization.
arXiv preprint arXiv:1911.08731
(2019)
\end{botherref}
\endbibitem

\bibitem[\protect\citeauthoryear{Blanchard et~al.}{2021}]{blanchard2021domain}
\begin{barticle}
\bauthor{\bsnm{Blanchard}, \binits{G.}},
\bauthor{\bsnm{Deshmukh}, \binits{A.A.}},
\bauthor{\bsnm{Dogan}, \binits{{\"U}.}},
\bauthor{\bsnm{Lee}, \binits{G.}},
\bauthor{\bsnm{Scott}, \binits{C.}}:
\batitle{Domain generalization by marginal transfer learning}.
\bjtitle{The Journal of Machine Learning Research}
\bvolume{22}(\bissue{1}),
\bfpage{46}--\blpage{100}
(\byear{2021})
\end{barticle}
\endbibitem

\bibitem[\protect\citeauthoryear{Krueger et~al.}{2021}]{krueger2021out}
\begin{bchapter}
\bauthor{\bsnm{Krueger}, \binits{D.}},
\bauthor{\bsnm{Caballero}, \binits{E.}},
\bauthor{\bsnm{Jacobsen}, \binits{J.-H.}},
\bauthor{\bsnm{Zhang}, \binits{A.}},
\bauthor{\bsnm{Binas}, \binits{J.}},
\bauthor{\bsnm{Zhang}, \binits{D.}},
\bauthor{\bsnm{Le~Priol}, \binits{R.}},
\bauthor{\bsnm{Courville}, \binits{A.}}:
\bctitle{Out-of-distribution generalization via risk extrapolation (rex)}.
In: \bbtitle{International Conference on Machine Learning},
pp. \bfpage{5815}--\blpage{5826}
(\byear{2021}).
\bcomment{PMLR}
\end{bchapter}
\endbibitem

\bibitem[\protect\citeauthoryear{Balaji et~al.}{2018}]{balaji2018metareg}
\begin{botherref}
\oauthor{\bsnm{Balaji}, \binits{Y.}},
\oauthor{\bsnm{Sankaranarayanan}, \binits{S.}},
\oauthor{\bsnm{Chellappa}, \binits{R.}}:
Metareg: Towards domain generalization using meta-regularization.
Advances in neural information processing systems
\textbf{31}
(2018)
\end{botherref}
\endbibitem

\bibitem[\protect\citeauthoryear{Qian et~al.}{2021}]{qian2021latent}
\begin{bchapter}
\bauthor{\bsnm{Qian}, \binits{H.}},
\bauthor{\bsnm{Pan}, \binits{S.J.}},
\bauthor{\bsnm{Miao}, \binits{C.}}:
\bctitle{Latent independent excitation for generalizable sensor-based
  cross-person activity recognition}.
In: \bbtitle{Proceedings of the AAAI Conference on Artificial Intelligence},
vol. \bseriesno{35},
pp. \bfpage{11921}--\blpage{11929}
(\byear{2021})
\end{bchapter}
\endbibitem

\bibitem[\protect\citeauthoryear{Sun and Saenko}{2016}]{sun2016deep}
\begin{bchapter}
\bauthor{\bsnm{Sun}, \binits{B.}},
\bauthor{\bsnm{Saenko}, \binits{K.}}:
\bctitle{Deep coral: Correlation alignment for deep domain adaptation}.
In: \bbtitle{European Conference on Computer Vision},
pp. \bfpage{443}--\blpage{450}
(\byear{2016}).
\bcomment{Springer}
\end{bchapter}
\endbibitem

\bibitem[\protect\citeauthoryear{Cremer et~al.}{2018}]{cremer2018inference}
\begin{bchapter}
\bauthor{\bsnm{Cremer}, \binits{C.}},
\bauthor{\bsnm{Li}, \binits{X.}},
\bauthor{\bsnm{Duvenaud}, \binits{D.}}:
\bctitle{Inference suboptimality in variational autoencoders}.
In: \bbtitle{International Conference on Machine Learning},
pp. \bfpage{1078}--\blpage{1086}
(\byear{2018}).
\bcomment{PMLR}
\end{bchapter}
\endbibitem

\bibitem[\protect\citeauthoryear{Krishnan
  et~al.}{2022}]{krishnan2022bayesiantorch}
\begin{botherref}
\oauthor{\bsnm{Krishnan}, \binits{R.}},
\oauthor{\bsnm{Esposito}, \binits{P.}},
\oauthor{\bsnm{Subedar}, \binits{M.}}:
Bayesian-Torch: Bayesian Neural Network Layers for Uncertainty Estimation.
\doiurl{10.5281/zenodo.5908307}
\end{botherref}
\endbibitem

\bibitem[\protect\citeauthoryear{Lin et~al.}{2017}]{lin2017focal}
\begin{bchapter}
\bauthor{\bsnm{Lin}, \binits{T.-Y.}},
\bauthor{\bsnm{Goyal}, \binits{P.}},
\bauthor{\bsnm{Girshick}, \binits{R.}},
\bauthor{\bsnm{He}, \binits{K.}},
\bauthor{\bsnm{Doll{\'a}r}, \binits{P.}}:
\bctitle{Focal loss for dense object detection}.
In: \bbtitle{Proceedings of the IEEE International Conference on Computer
  Vision},
pp. \bfpage{2980}--\blpage{2988}
(\byear{2017})
\end{bchapter}
\endbibitem

\end{thebibliography}
\end{document}